\pdfoutput=1

\documentclass[11pt]{article}

\usepackage[final]{acl}

\usepackage{times}
\usepackage{latexsym}
\usepackage{graphicx}

\usepackage[T1]{fontenc}

\usepackage[utf8]{inputenc}

\usepackage{microtype}
\usepackage{amsmath}
\usepackage{amssymb}

\usepackage{booktabs}
\usepackage{multirow}
\usepackage{adjustbox}
\usepackage{url}
\usepackage{enumitem}
\usepackage{etoolbox,setspace} 
\AtBeginEnvironment{quote}{\par\singlespacing\small}
\usepackage{listings}
\usepackage{xcolor}

\title{Dialogue Summaries as Dialogue States (DS2),\\
Template-Guided Summarization for Few-shot Dialogue State Tracking}

\author{
Jamin Shin$^{1}$\thanks{\quad Equal Contribution: JS proposed the main idea and scaled up the experiments. HY designed and implemented the heuristic state tracking component. HM conducted rapid prototyping,  analysis, and ablations.}$^*$,
Hangyeol Yu$^{1*}$,
Hyeongdon Moon$^{1*}$,
Andrea Madotto$^{2}$,
Juneyoung Park$^1$\\ 
$^1$Riiid AI Research\\
$^2$The Hong Kong University of Science and Technology \\
\texttt{jayshin.nlp@gmail.com},
\texttt{\{hangyeol.yu,hyeongdon.moon\}@riiid.co}\\
\texttt{amadotto@connect.ust.hk},
\texttt{juneyoung.park@riiid.co}
}

\newcommand{\CT}[0]{\texttt{CT}}
\newcommand{\CD}[0]{\texttt{CD}}
\newcommand{\MD}[0]{\texttt{MD}}
\begin{document}
\maketitle

\begin{abstract}
Annotating task-oriented dialogues is notorious for the expensive and difficult data collection process. 
Few-shot dialogue state tracking (DST) is a realistic solution to this problem.
In this paper, we hypothesize that dialogue summaries are essentially unstructured dialogue states; hence, we propose to reformulate dialogue state tracking as a dialogue summarization problem.
To elaborate, we train a text-to-text language model with synthetic template-based dialogue summaries, generated by a set of rules.
Then, the dialogue states can be recovered by inversely applying the summary generation rules.
We empirically show that our method DS2 outperforms previous works on few-shot DST in MultiWoZ 2.0 and 2.1, in both cross-domain and multi-domain settings.
Our method\footnote{Code: \href{https://github.com/jshin49/ds2}{\nolinkurl{github.com/jshin49/ds2}}} also exhibits vast speedup during both training and inference as it can generate all states at once.
Finally, based on our analysis, we discover that the naturalness of the summary templates plays a key role for successful training.

\end{abstract}


\section{Introduction}

\begin{figure}[ht]
  \includegraphics[width=0.45\textwidth]{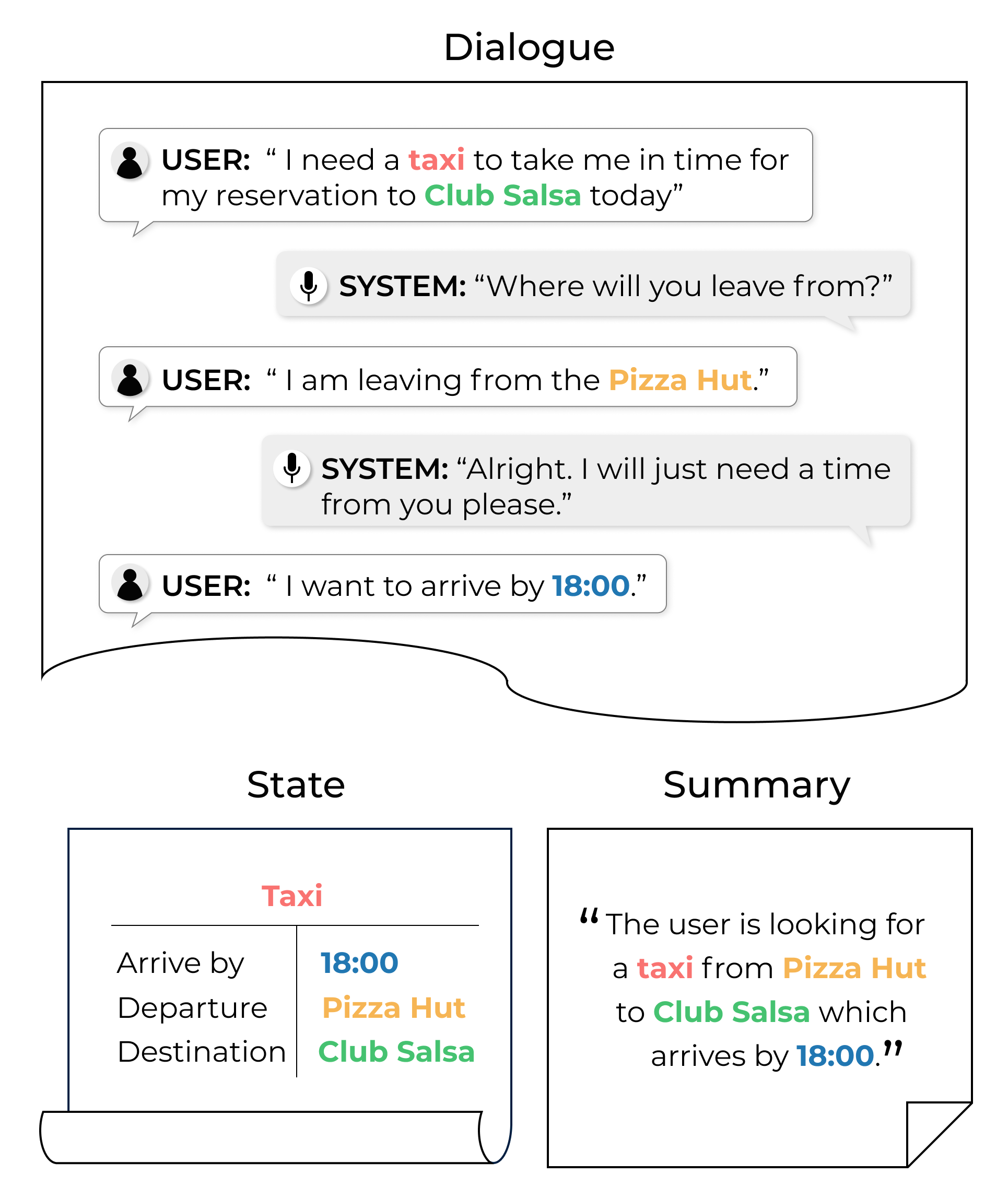}
    \caption{Example dialogue in taxi domain, its dialogue state, and template summary created from the state.}
  \label{fig:intro}
\end{figure}

\begin{figure*}[ht]
\includegraphics[width=\textwidth]{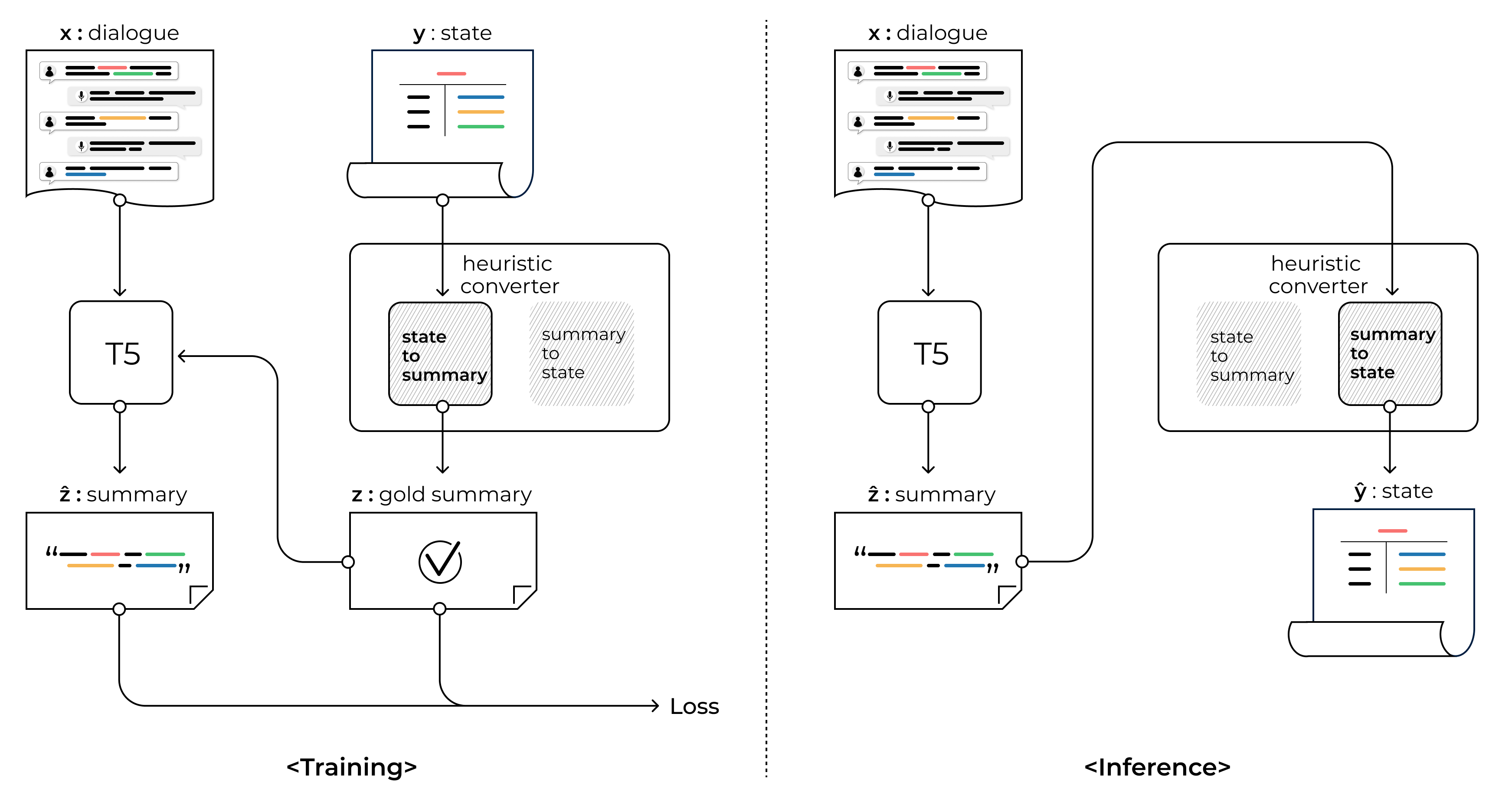}
\caption{Overall picture of our method DS2.}
\label{fig:training-inference}
\end{figure*}

Task-oriented dialogue systems (TOD) have penetrated our daily lives much more than before and they will continue to increase their presence. 
For example, many of our mobile devices are equipped with such dialogue agents like Siri, and we now often encounter customer service or flight reservation bots.
Dialogue State Tracking (DST) is an essential element of such task-oriented dialogue systems~\cite{wu2019transferable,balaraman2021recent}.
The main goal of this component is to understand the user's requirements expressed during the conversation under a given schema or ontology. 
Hence, as shown in Figure~\ref{fig:intro}, accurately extracting the departure, destination, and arrival time of the user is key to creating a good user experience.

However, collecting such turn-level dialogue state annotations is very expensive and requires a significant amount of design and mediating efforts from domain experts~\cite{budzianowski2018large,eric-etal-2020-multiwoz,park2021klue}. This is because the collection process follows the Wizard-of-Oz (WoZ) style~\cite{kelley1984wozsetting}, which requires two human workers to converse with each other and annotate the states for each turn. To cope with this inherent scalability issue,~\citet{budzianowski2018large} attempted to crowd-source this process in MultiWoZ 2.0 which resulted in one of the largest publicly available multi-domain task-oriented dialogue dataset. 
However, the resulting dataset is very noisy in data annotations which often hindered the training and evaluation process. In fact, the community has already seen 4 different revisions of this dataset from 2.1 to 2.4~\cite{eric-etal-2020-multiwoz,zang2020multiwoz22,han2020multiwoz23,ye2021multiwoz24}.

Furthermore, in realistic industrial settings, having to expand the existing model and ontology to include new domains and slot-values is a common phenomenon. Naturally, there have been many recent works that proposed (zero) few-shot settings to rely on less annotated data. For instance, both STARC~\cite{gao2020starc} and TransferQA~\cite{lin2021zero} achieve great few-shot DST performance on MultiWoZ 2.0 by prompting large pre-trained language models like BERT~\cite{devlin2019bert} and T5~\cite{raffel2020t5} with natural language questions (e.g. ``In what area is the user looking for a hotel?'').

Meanwhile, despite the good performance of the aforementioned works, they still suffer from certain issues. \textbf{1) They often require} a large amount of expensive labeled training data from other tasks or domains for task-specific pre-training. For example, as shown in Table~\ref{tab:used_pretrained_data}, SOLOIST~\cite{peng2020soloist} uses $\sim766K$, TOD-BERT~\cite{wu2020tod} uses $\sim1.39M$, and PPTOD~\cite{su2021multi} utilizes $\sim2.37M$ dialogue utterances. Meanwhile, TransferQA~\cite{lin2021zero} also uses a vast amount of QA data ($\sim$720K).
\textbf{2) QA-style prompting} as in TransferQA~\cite{lin2021zero} not only requires additional efforts to handle ``none'' and ``yes-no'' slots
but also has an expensive slot-value decoding time complexity: $k$ times inference of a language model where $k$ is the number of slots. Overall, the aforementioned works are still expensive in terms of time, money, and engineering costs.




Addressing the above challenge, we propose to cast Dialogue State Tracking as a Dialogue Summarization task; hence the name is \textbf{Dialogue Summaries as Dialogue States (DS2)}. 
The main hypothesis for this reformulation is that \textit{dialogue summaries are essentially unstructured dialogue states}.
In this paper, we explore this reformulation to the limit. We fine-tune large text-to-text pre-trained language models (e.g. T5, BART) with synthetic dialogue summaries. These summaries are created by heuristic rules from the dialogue states, as in Figure~\ref{fig:intro}.
Hence, as these models already excel in text summarization, the research question we ask is \textit{whether we can guide dialogue summarization models to generate dialogue summaries that conform to the templates we provide}.
Then, we can extract the dialogue states by inversely applying the rules we used to create the synthetic summaries.

Compared to previous approaches, our method has several advantages that come naturally. 
First, we easily reduce the pre-train \& fine-tune discrepancies without any DST-specific engineering by leveraging dialogue summarization datasets.
These datasets are an order of magnitude smaller in annotated data size (e.g. SAMSum~\cite{gliwa2019samsum} has $\sim200K$ utterances). 
Second, we achieve great speedup in both training and inference because we only need to summarize once, and we can extract slot values from the summary with negligible cost.

Finally, the significant improvement that DS2 brings to MultiWoZ 2.0 and 2.1 datasets in the few-shot DST performance for both \textit{cross-domain} and \textit{multi-domain} settings empirically show the effectiveness of our approach. \textit{Without extensively using such expensive annotated data for pre-training}, DS2 generally outperforms previous works that do so. In our analysis, we also show how naturalness of the summary has played a key role for this work.
Our main contribution can be summarized as such:
\begin{itemize}
    \item We propose DS2, which is the first approach to cast Dialogue State Tracking as Dialogue Summarization.
    \item Our formulation provides relatively easier reduction of pre-train \& fine-tune discrepancy, while also significantly improving training and inference speed for generative DST.
    \item We empirically show that our method outperforms previous methods in MultiWoZ 2.0 and 2.1 for both cross-domain and multi-domain few-shot DST settings.
\end{itemize}

\begin{table}[t]
\centering
\begin{adjustbox}{width=0.48\textwidth}
\begin{tabular}{@{}lll@{}}
\toprule
Model       & \# of Pre-train Data   & Data Type                        \\ \midrule
TOD-BERT\cite{wu2020tod}    & $\sim$1.39M & Dialogue Utterances \\
PPTOD \cite{su2021multi}      & $\sim$2.37M & Dialogue Utterances \\
Transfer QA \cite{lin2021zero}& $\sim$720K  & QA Pairs \\
SOLOIST   \cite{peng2020soloist}  & $\sim$766K  & Dialogue Utterances \\
Ours - DS2       & $\sim$199K  & Dialogue Utterances \\
\bottomrule
\end{tabular}
\end{adjustbox}
\caption{Pre-train data usage scale comparison with other models. We used SAMSum~\cite{gliwa2019samsum}, which is a dialogue summarization dataset, and we estimated the number of utterances in SAMSum to be in the range (154k, 243k).}
\label{tab:used_pretrained_data}
\end{table}

\section{Related Work}

\textbf{Dialogue State Tracking} is a well-known sub-task of task oriented dialog systems~\cite{williams2007partially,williams2014dialog}. The current state-of-the-art techniques fine-tune pre-trained language models~\cite{lei2018sequicity,zhang2020task,wu2020tod,peng2020soloist,zhang2020find,kim-etal-2020-efficient,lin2020mintl,Chen2020SchemaGuidedMD,heck2020trippy,mehri2020dialoglue,NEURIPS2020_e9462095,yu2020score,li2020coco} are often further trained with a large amount of annotated data. 

\textbf{Few-Shot DST} is a promising direction for reducing the need of human annotation while achieving quasi-SOTA performance with a fraction of the training data. Different techniques have been proposed
~\cite{wu2019transferable,mi2021cins,li2021zero,gao2020starc,lin2021t5dst,lin2021zero,campagna2020zero,wu2020improving,su2021multi,peng2020soloist,wu2020tod}. We briefly describe and compare DS2 with existing few-shot models in Section~\ref{sec:baselines}. 



\textbf{Dialogue Summarization}
The community has been seeing an increasing amount of interest in this subfield: from datasets~\cite{zhu2021mediasum,zhong2021qmsum,chen2021dialogsum,fabbri-etal-2021-convosumm,zhang2021emailsum} to models~\cite{wu2021cods,feng2021language,khalifa2021bag,chen2020multi}. 

\textbf{Prompt Engineering}
Many recent works on prompt engineering or Pattern Exploiting Training (PET)~\cite{schick2020few,schick2021exploiting,schick2021s,gao-etal-2021-making,liu2021pre,madotto2021few,ConstrainedLMSemanticParser2021,liu2021pre} have been proposed to explore prompt-based few-shot learning capabilities for Pre-trained Language Models. Interestingly, they share similar insights about the critical role of natural templates for successful few-shot learning. 

\section{Methodology}
\subsection{Background}
A data point for DST is a pair of a task-oriented dialogue $\mathbf{x}$ and a sequence $\{\mathbf{y}_t\}_{t=1}^{n}$ of dialogue states where $t$ and $n$ refer to current turn index and the total number of turns in the dialogue respectively. Here, $\mathbf{y}_t$ denotes the dialogue state after turn $t$. A dialogue state is a set of slot-value pairs, 
\[
\mathbf{y}_t = \{(k_1, v_1), (k_2, v_2), \ldots, (k_m, v_m)\}
\]
where the set of all possible slots $k_i$'s in a domain is predefined. For example, the attraction domain in MultiWoZ has three kinds of slots, namely, `attraction-area', `attraction-name', and `attraction-type'. 
With this setting, DST is a task to predict $\mathbf{y}_t$ given the truncated dialogue $\mathbf{x}_{1:t}$ as input for every $t$. For convenience, we will omit the turn index $t$.

\subsection{Overview: Dialogue Summaries as Dialogue States (DS2)}
\label{sec:methodology_overview}
In this section, we describe the overall picture of the proposed method, DS2. First, our method is composed of 3 components: the Pre-trained text-to-text Language Model (PLM; $\theta$) such as T5, dialogue summary generator (\textit{state-to-summary}; $\phi$), and dialogue state extractor (\textit{summary-to-state}; $\eta$). 
To briefly describe the training process, given a dialogue $\mathbf{x}$, we first generate a synthetic summary $\mathbf{z} = \phi(\mathbf{y})$ as in Table~\ref{tab:slot_template}, using the \textit{state-to-summary} module. 
Instead of generating dialogue states directly as done by~\citet{wu2019transferable,gao2019dstreader}, we fine-tune the PLM to predict $\mathbf{z}$.
The training loss is then calculated between $\mathbf{z}$ and $\hat{\mathbf{z}}$, which is the cross-entropy loss between $P(\mathbf{z}~|~\mathbf{x})$ and the predicted summary $\mathbf{z}$.
This process is described in the <Training> part of Figure~\ref{fig:training-inference} (left section). Note that the only module that we train is the summary model $\theta$. 
During inference, the PLM generates a summary $\hat{\mathbf{z}}$ and the dialogue state $\hat{\mathbf{y}}$ is extracted from it using the \textit{summary-to-state} module $\eta$. The right section of Figure~\ref{fig:training-inference} describes this process.



Our method DS2 reformulates DST into a summarization task. The idea is simple. If a model summarizes a given dialogue with all the slot-value information, \textit{exactly in the format we want}, then we can simply use regular expressions to parse the slot-values from the generated summary. The mathematical assumption here is that the \textit{state-to-summary} converter $\phi$ is a left inverse of the \textit{summary-to-state} converter $\eta$. That is, $\eta(\phi(\mathbf{y}')) = \mathbf{y}'$ for every dialogue state $\mathbf{y}'$. Let $(\mathbf{x}, \mathbf{y})$ be a training sample. If a predicted summary $\hat{\mathbf{z}} = \theta(\mathbf{x})$ exactly matches the generated one $\mathbf{z} = \phi(\mathbf{y})$, the later step is straight forward by $\eta$ as follows:
\[
\eta (\theta(\mathbf{x})) = \eta(\hat{\mathbf{z}}) = \eta(\mathbf{z}) = \eta(\phi(\mathbf{y})) = \mathbf{y}.
\]
Here, $\eta \circ \theta$ is the DST model we want. 

Note that the space of all texts is larger than the set of all dialogue states defined by the ontology. The former is infinite but the latter is finite. Therefore, there is no one-to-one correspondence between two sets. That is one reason we consider a certain template for summaries: to restrict the set of candidate summaries so that the size perfectly matches the set of all states. One more benefit from the template is that it naturally provides a structural summary-to-state conversion. 

Meanwhile, the reduced summarization task is subtle because, at least, a generated summary $\theta(\mathbf{x})$ must satisfy the template to guarantee our argument. In mathematical words, $\theta(\mathbf{x})$ should be in the image of $\phi$. In general, it is nontrivial to control a deep learning model so that its output is always in an arbitrary subset, and it is even harder with few samples. Therefore, we hypothesize that the naturalness of the template is a key factor to the performance of our model.


\begin{table}[t!]
\centering
\begin{adjustbox}{width=0.45\textwidth}
\begin{tabular}{@{}ccc@{}}
\toprule
\textbf{Slot Name} & \textbf{Slot Template} & \textbf{Slot Value} \\ \midrule
attraction-area    & located in the \_      & \textit{center}     \\
attraction-name    & called \_              & \textit{byard art}  \\
attraction-type    & which is a \_          & \textit{museum}     \\ \midrule\midrule
\multirow{2}{*}{\textbf{\begin{tabular}[c]{@{}c@{}}Sentence\\ Prefix\end{tabular}}} &
  \multicolumn{2}{c}{\multirow{2}{*}{The user is looking for an attraction}} \\
                   & \multicolumn{2}{c}{}                         \\ \midrule
\multirow{3}{*}{\textbf{\begin{tabular}[c]{@{}c@{}}Example\\ Synthetic\\ Summary\end{tabular}}} &
  \multicolumn{2}{c}{\multirow{3}{*}{\begin{tabular}[c]{@{}c@{}}``The user is looking for an attraction\\ called \emph{byard art} which is a \emph{museum}\\ located in the \emph{center}.''\end{tabular}}} \\
                   & \multicolumn{2}{c}{}                         \\
                   & \multicolumn{2}{c}{}                         \\ \bottomrule
\end{tabular}
\end{adjustbox}
\caption{Template for attraction domain in MultiWoZ.
}
\label{tab:slot_template}
\end{table}

\subsection{State-to-summary Converter}
\label{sec:state-to-sum}
For each dialogue domain, we manually wrote a template to automatically synthesize human-readable summaries from dialogue states. Designing the template, domain-specific information is considered such as the name of slots and possible values. Table~\ref{tab:slot_template} illustrates a template for the ``attraction'' domain in MultiWoZ with example values. This template itself can be regarded as the previously discussed function $\phi$, which takes a dialogue state as an input to produce a dialogue summary. 

Given a state, the corresponding summary is built based on the template in a hierarchical manner. Suppose there are $m$ slots in the current domain, namely, $k_1, \ldots, k_m$. We define a phrase template $p_i$ for each slot $k_i$, which is a function that takes a value string as input and produces a phrase. In Table~\ref{tab:slot_template}, the slot named ``attraction-area'' is mapped to a phrase template ``located in the \_''. After combining with the slot-value \emph{centre}, we get a phrase ``located in the \emph{centre}''.
Let $\mathbf{y} = \{(k_1, v_1), \ldots, (k_{m'}, v_{m'})\}$ be a given state where $m' \leq m$. Each value $v_i$ of a slot appearing in the state is matched to the phrase template $p_i$, so we get the set of phrases $\{p_1(v_1), \ldots, p_{m'}(v_{m'})\}$. They are joined together and added to the sentence prefix of the domain such as ``The user is looking for an attraction'', to get the final summary: 

\begin{quote}
``The user is looking for an attraction called \emph{byard art} which is a \emph{museum} located in the \emph{centre}.''
\end{quote}

The template also covers exceptional cases, \emph{dontcare}. Each slot has a special phrase for \emph{dontcare}. For example, ``attraction-area'' is mapped to a phrase ``the location''. For that case, another sentence prefix ``, and he does not care about'' is used. The resulting summary is:

\begin{quote}
``The user is looking for an attraction which is a \emph{museum}, and he does not care about \emph{the location}.''
\end{quote}

We do not care too much about \emph{none} values as it is naturally covered. Since we remove all slots whose values are \emph{none}, following our \textit{state-to-summary} converting method, the synthesized gold summary does not mention those slots. This behavior conforms to commonsense such that a summary generally does not include information not mentioned in the source text.

In a MultiWoZ dialogue, speakers often talk about multiple domains, so the synthesized summary should also mention the values from multiple domains. Given a multi-domain state, we split the state by different domains, and convert each single-domain partial state to a summary sentence. Then the resulting sentences are connected to a multiple-sentence summary. To be more natural, we paraphrased the common sentence prefix ``The user is looking for'' to ``He is searching for'' or ``He looks for'' for later utterances.
For more examples, please refer to the Appendix Table~\ref{tab:domain_template_ex}.


\subsection{Summary-to-state Converter}
From a generated summary, a dialogue state is extracted by summary-to-state converter $\eta$. Based on the same template, the process is almost\footnote{some slot-value entities include prepositions.} the inverse of summary synthesis. We first split the whole summary into sentences from different domains. Domain-specific sentence prefix is used to identify which sentence is from which domain. The remaining process is to convert each single domain's one-sentence summary to a single-domain dialogue state and to finally merge them into one set of states. 
To convert a single-domain summary, slot values are extracted through string pattern matching via regular expressions based on the slot phrase templates from Section~\ref{sec:state-to-sum}. 

\begin{table*}[ht!]
\begin{adjustbox}{width=1\textwidth}
\begin{tabular}{cccccccccccccccc}
\hline
\multirow{2}{*}{Model (\textit{ver.} / \texttt{mode})} & \multicolumn{3}{c}{Attraction}                         & \multicolumn{3}{c}{Hotel}                              & \multicolumn{3}{c}{Restaurant}                      & \multicolumn{3}{c}{Taxi}                            & \multicolumn{3}{c}{Train}                              \\ \cline{2-16} 
                                                       & 1\%            & 5\%               & 10\%              & 1\%               & 5\%               & 10\%           & 1\%            & 5\%            & 10\%              & 1\%               & 5\%            & 10\%           & 1\%            & 5\%               & 10\%              \\ \hline
TRADE (2.0 / \CD)                              & 35.8           & 57.5              & 63.1              & 19.7              & 37.4              & 41.4           & 42.4           & 55.7           & 60.0              & 63.8              & 66.5           & 70.1           & 59.8           & 69.2              & 71.1              \\
DSTQA (2.0 / \CD)                              & -              & \textbf{70.4}     & \textbf{71.6}     & -                 & 50.1              & 53.6           & -              & 58.9           & 64.5              & -                 & 70.9           & 74.1           & -              & 70.3              & 74.5              \\
T5-DST (2.0 / \CD)                             & 58.8           & 65.7              & 69.5              & 43.1              & 50.7              & 54.9           & 57.6           & 61.9           & 63.5              & 70.1              & 73.7           & 74.7           & 70.8           & 74.2              & 77.6              \\
CINS (2.0 / \CT)                               & 45.6           & 61.2              & -                 & 33.9              & 46.2              & -              & 40.6           & 53.9           & -                 & 59.7              & 63.3           & -              & 60.3           & 73.8              & -                 \\
STARC (2.0 / \CT)                              & 40.3           & 65.3              & 66.2              & \textbf{45.9}     & \textbf{52.5}     & \textbf{57.3}  & 51.6           & 60.4           & 64.6              & 72.5              & 75.3           & 79.6           & 65.6           & 74.1              & 75.0              \\
TransferQA (2.0 / \CT)                         & 52.3           & 63.5              & 68.2              & 43.4              & 52.1              & 55.7           & 51.7           & 60.7           & 62.9              & \textbf{75.4}     & \textbf{79.2}  & \textbf{80.3}  & 70.1           & 75.6              & \textbf{79.0}     \\ \hline
DS2 (2.0 / \CD)                                & \textbf{65.26} & \underline{69.40} & \underline{70.89} & \underline{44.34} & \underline{52.16} & 53.79          & \textbf{58.94} & \textbf{64.12} & \textbf{64.65}    & \underline{74.15} & 77.18          & 78.50          & \textbf{74.21} & \textbf{76.96}    & \underline{78.60} \\
DS2 (2.0 / \CT)                                & 55.84          & 65.32             & 68.73             & 37.78             & 48.02             & 51.82          & 48.57          & 61.37          & 64.61             & 68.62             & 72.60          & 75.53          & 70.37          & \underline{75.68} & \underline{78.16} \\
DS2 (2.0 / \MD)                                & 62.28          & \underline{69.30} & \underline{70.88} & 38.65             & 50.61             & 51.20          & 54.46          & 61.98          & \underline{64.52} & 71.03             & 75.10          & 76.90          & 70.41          & \underline{75.87} & \underline{78.08} \\ \midrule\midrule
TransferQA (2.1 / \CT)                         & 50.25          & 60.92             & 64.28             & 32.46             & 39.02             & 41.99          & 47.12          & 59.16          & 62.24             & 71.12             & 74.47          & 76.07          & 69.01          & 73.17             & 75.46             \\ \hline
DS2 (2.1 / \CD)                                & \textbf{60.04} & \textbf{68.74}    & \textbf{70.31}    & \textbf{43.02}    & \textbf{48.44}    & \textbf{50.35} & \textbf{56.54} & \textbf{65.11} & \textbf{67.26}    & \textbf{76.41}    & \textbf{79.81} & \textbf{80.62} & \textbf{73.07} & \textbf{76.18}    & 77.00             \\
DS2 (2.1 / \CT)                                & 53.60          & 64.44             & 66.90             & 36.17             & 46.96             & 48.29          & 48.36          & 63.96          & 66.82             & 68.84             & 76.82          & 77.23          & 67.96          & 75.55             & \textbf{77.14}    \\
DS2 (2.1 / \MD)                                & 56.33          & 66.39             & 67.14             & 38.22             & 47.75             & 48.34          & 50.19          & 63.22          & 64.45             & 71.87             & 77.10          & 79.01          & 69.87          & 75.55             & 76.36             \\ \hline
\end{tabular}
\end{adjustbox}
\caption{Per-domain few-shot (1-5-10\%) results on MultiWOZ 2.0 and 2.1 (\textit{ver.}). All of our \textbf{DS2} results are averaged over \textit{3 runs (seeds)} and full results of each run are in the Appendix Tables~\ref{tab:all_run_jgas},\ref{tab:bart_performance}. \CD, \CT, \MD~ each refer to \textit{Cross-Domain}, \textit{Cross-Task}, \textit{Multi-Domain} few-shot scenarios. We pre-trained TransferQA ourselves and fine-tuned it on \textit{ver.} 2.1 to get the results, while all other results were taken from their respective papers. Note that we compare \CD, \CT, \MD~ together as they all share the same test-set. Our proposed model \textbf{DS2} based on \texttt{T5-large} either achieves \textbf{SOTA} (bold) or \underline{competitive} (underlined; $\sim$1.5-point difference) results in 2.0, and for 2.1 with the \CD~ setting we outperform the SOTA model in 2.0 - TransferQA.}
\label{tab:cd_jgas}
\end{table*}

\section{Experiments}

\subsection{Dataset}
\label{subsec:dataset}

\textbf{MultiWoZ}~\cite{budzianowski2018large} is a large-scale English multi-domain task-oriented dialogue dataset. It contains 7 different domains, but as in \citet{wu2019transferable}, we only use 5 out of them: train, hotel, restaurant, attraction, and taxi. Table~\ref{tab:dialogue_count} shows the number of dialogues for each domain in the training set of MultiWoZ 2.1.
We evaluate DS2 on both MultiWoZ 2.0 and MultiWoZ 2.1, as most of the benchmark performances were reported on MultiWoZ 2.0. 

\noindent\textbf{SAMSum}~\cite{gliwa2019samsum} is a dialogue summarization dataset. We further pre-train T5-large~\cite{raffel2020t5} more with SAMSum by using the code from \citet{wu2021cods} before we fine-tune for DS2. 

\subsection{Evaluation}

\textbf{DST}
The main performance metric for our few-shot DST experiments is Joint Goal Accuracy (JGA). 
For each turn, only if the model's output dialogue state is exactly the same as the set of gold labels, we consider it correct~\cite{balaraman2021recent}. We report both all-domain JGA and per-domain JGA as in \citet{wu2019transferable} based on the evaluation setting that is described in the below Section~\ref{sec:few-shot-settings}. Slot accuracy is also computed for both active slots and none slots.

\noindent\textbf{Dialogue Summarization}
In addition to the metrics for dialogue state prediction, we also use metrics to measure the quality of the intermediate dialogue summaries $\hat{\mathbf{z}}$. We measure BLEU-4 \cite{papineni2002bleu} and ROUGE-4 (F1) \cite{lin-2004-rouge} scores to evaluate how close a model-generated summary is to the synthesized gold summary. We also use ROUGE score to measure the performance of pre-training T5-large on the SAMSum corpus. The summarization performance are shown in Table~\ref{tab:dial_summ}.

\subsection{Model}
\label{sec:model}
We mainly experiment with two pre-trained language models, T5-large and BART-large, as the summarization models of DS2. 
The pre-trained weights of T5-large from \citet{raffel2020t5} are trained on mail and news data summarization. 
Hence, as mentioned above, we further pre-train the model with dialogue summarization.\footnote{The T5-large model we pre-trained on SAMSum corpus is released here: \url{https://huggingface.co/jaynlp/t5-large-samsum}} 
To be specific, we pre-pend the prefix \textit{Summarize this dialogue:} to $\mathbf{x}$ as done in the recent T0~\cite{sanh2021t0}. 
We use BART-large that is already pre-trained on both XSum~\cite{narayan2018xsum} and SAMSum from~\citet{wu2021cods}. 
In the ablation studies (Section~\ref{subsec:ablation_study}), to compare the effectiveness of SAMSum pre-training, we use the original BART-large pre-trained on XSum~\cite{lewis2020bart}. 

\begin{table}[t]
\centering
\begin{adjustbox}{width=0.4\textwidth}
\begin{tabular}{@{}ccc@{}}
\toprule
MultiWoZ 2.1 & single-domain & multi-domain \\ \midrule
Hotel        & 513           & 3381         \\
Taxi         & 325           & 1654         \\
Attraction   & 127           & 2717         \\
Restaurant   & 1197          & 3813         \\
Train        & 275           & 3103         \\ \bottomrule
\end{tabular}
\end{adjustbox}
\caption{Number of dialogues for each domain in MultiWoZ 2.1 training set. Single-domain dialogues are subset of multi-domain dialogues.}
\label{tab:dialogue_count}
\end{table}

\begin{table}[t]
\centering
\begin{adjustbox}{width=0.48\textwidth}
\begin{tabular}{@{}ccccc@{}}
\toprule
Model                                                                     & Rouge-1        & Rouge-2        & Rouge-L        & \# Params \\ \midrule
\begin{tabular}[c]{@{}c@{}}PEGASUS\\ \cite{zhang2020pegasus}\end{tabular} & 50.50          & 27.23          & 49.32          & ~568M     \\
\begin{tabular}[c]{@{}c@{}}BART-large\\ \cite{lewis2020bart}\end{tabular} & 51.74          & 26.46          & 48.72          & ~406M     \\
\begin{tabular}[c]{@{}c@{}}T5-large\\ \cite{raffel2020t5}\end{tabular}    & \textbf{52.69} & \textbf{27.42} & \textbf{49.85} & ~770M     \\ \bottomrule
\end{tabular}
\end{adjustbox}
\caption{Dialogue summarization results on SAMSum corpus~\cite{gliwa2019samsum}. Both BART and PEGASUS numbers are taken from~\citet{wu2021cods}, while for T5-large, we pretrained it using the code from~\citet{wu2021cods}. Given such summarization results, we choose to use T5-large and BART-large.}
\label{tab:dial_summ}
\end{table}

\begin{table*}[h!t]
\centering
\begin{adjustbox}{width=0.9\textwidth}
\begin{tabular}{@{}lcccc@{}}
\toprule
\multicolumn{1}{c}{Model (\textit{ver.})}             & 1\%                   & 5\%                   & 10\%                  & 100\% \\ \midrule
TRADE (2.0)~\cite{wu2019transferable}                 & 11.74 (-)             & 32.41 (-)             & 37.42 (-)             & 48.62 \\
TRADE + Self-supervision (2.0)~\cite{wu2020improving} & 23.0 (-)              & 37.82 (-)             & 40.65 (-)             & -     \\
MinTL* (2.0)~\cite{lin2020mintl}                      & 9.25 (2.33)           & 21.28 (1.94)          & 30.32 (2.14)          & 52.10 \\
SOLOIST* (2.0)~\cite{peng2020soloist}                 & 13.21 (1.97)          & 26.53 (1.62)          & 32.42 (1.13)          & 53.20 \\
PPTOD* (2.0)~\cite{su2021multi}                       & 31.46 (0.41)          & 43.61 (0.42)          & 45.96 (0.66)          & 53.89 \\ \midrule
DS2 - T5 (2.0)                                        & \textbf{36.15 (1.87)} & \textbf{45.14 (1.69)} & \textbf{47.61 (0.37)} & 54.78 \\ \midrule\midrule
TRADE (2.1)~\cite{wu2020improving}                    & 12.58 (-)             & 31.17 (-)             & 36.18 (-)             & 46.00 \\
TRADE + Self-supervision (2.1)~\cite{wu2020improving} & 21.90 (-)             & 35.13 (-)             & 38.12 (-)             & -     \\ \midrule
DS2 - BART (2.1)                                      & 28.25 (0.98)          & 37.71 (1.05)          & 40.29 (0.29)          & 46.86 \\
DS2 - T5 (2.1)                                        & \textbf{33.76 (1.49)} & \textbf{44.20 (0.98)} & \textbf{45.38 (1.05)} & 52.32 \\ \bottomrule
\end{tabular}
\end{adjustbox}
\caption{Multi-domain Few-shot (1-5-10\%) JGA evaluated on all domains jointly. *: taken from PPTOD~\cite{su2021multi}. Our models were run 3 times and full results are in Appendix Table~\ref{tab:md_all_performance}.}
\label{tab:md_jgas}
\end{table*}

\subsection{Few-Shot Settings}
\label{sec:few-shot-settings}
There are three different scenarios for few-shot DST experiments:
\begin{itemize}
    \item Cross-Domain (\CD)~\cite{wu2019transferable}
    \item Cross-Task (\CT)~\cite{gao2020starc}
    \item Multi-Domain (\MD)~\cite{wu2020improving}
\end{itemize}
For each setting, \textbf{1\%, 5\%, 10\%}, or \textbf{100\%} of training data is sampled to fine-tune a model. 
For all settings, we use the entire dev and test data for evaluation. 
As described in Section~\ref{subsec:dataset}, we run each scenario for both MultiWoZ 2.0 and 2.1.

\paragraph{Cross-Domain} \CD~was first explored by~\citet{wu2019transferable} in MultiWoZ 2.0. 
In this setting, we consider the scenario of adapting a Dialogue System to a new target domain (e.g. taxi) while we have full training data for the source domains (e.g. restaurant, hotel, attraction, train). 
For this setting, we pre-train DS2 on all the source domains and then fine-tune the target domain.
Note that during target-domain fine-tuning, as most of the dialogues are multi-domain (Table~\ref{tab:dialogue_count}), we train DS2 to output summaries for \textbf{all domains} during the adaptation as well. 
During evaluation, only per-domain JGA is reported as in~\citet{wu2019transferable}. 

\paragraph{Cross-Task} \CT~ was first explored for MultiWoZ by~\citet{gao2020starc} to demonstrate zero-shot DST performance. 
In our case, the difference with \CD~ is that there is no source-domain pre-training and only target-domain fine-tuning is done.
We measure per-domain JGA exactly as we do in \CD.

\paragraph{Multi-Domain} For ~\MD~ experiments all domains are used to train a model. Every slot value is used for both summary synthesis and evaluation. Both JGA per-domain and total JGA are measured for multi-domain DST. We also evaluate full-shot training for multi-domain DST.

\subsection{Baselines}
\label{sec:baselines}

All baseline results are only reported on MultiWoZ 2.0, and we additionally experimented with TransferQA on 2.1 as it was the best model.

\paragraph{TRADE} (\CD, \MD)~\citet{wu2019transferable} utilizes copy mechanism and slot \& domain embeddings for transferability. Meanwhile, \citet{wu2020improving} applies self-supervision to improve zero-shot and few-shot \CD \& \MD~ performances of TRADE.

\paragraph{T5-DST} (\CD)~\citet{lin2021t5dst} prompts a T5 model with slot descriptions for few-shot DST.

\paragraph{STARC} (\CT)~\citet{gao2020starc} asks natural language questions separately to two different instances of RoBERTa-Large~\cite{liu2019roberta} for categorical and non-categorical slots.

\paragraph{TransferQA} (\CT)~\citet{lin2021zero} asks natural language questions to a single T5-large model that is pre-trained to predict none values properly. As the original authors did not release their pre-trained version we release our own using their code\footnote{TransferQA pre-trained on the QA data: \url{https://huggingface.co/jaynlp/t5-large-transferqa}}.

\paragraph{CINS} (\CT)~\citet{mi2021cins} prompts a T5-base with slot descriptions for few-shot DST. 

\paragraph{DSTQA} (\CD)~\citet{Zhou2019MultidomainDS} performs DST via question answering over ontology graph.



\paragraph{PPTOD} (\MD)~\cite{su2021multi} prompts a PLM pre-trained on various TOD task and data with natural language instructions.

\section{Result}
\label{sec:results}

\begin{table*}[ht]
\centering
\begin{adjustbox}{width=0.95\textwidth}
\begin{tabular}{@{}ll@{}}
\toprule
Error type & \textbf{Hallucination} : The model generates unmentioned information.                                                                                                \\ \midrule
Pattern    & The user is looking for a train from \_\_\_\_ to \_\_\_\_ on \_\_\_\_, which leaves at \_\_\_\_.                      \\
Summary    & The user is looking for a train \textcolor{red}{for 7 people} from \textcolor{blue}{broxbourne} to \textcolor{blue}{cambridge} on \textcolor{blue}{wednesday}, which arrives at \textcolor{blue}{11:30}.       \\
Gold   & The user is looking for a train from broxbourne to cambridge on wednesday, which leaves at 11:30.                     \\ \midrule
Error type & \textbf{Missing slot} : The model omits expected slot.                                                                                                 \\ \midrule
Pattern    & The user is looking for a train from \_\_\_\_ on \_\_\_\_, which leaves at \_\_\_\_.                                  \\
Summary    & The user is looking for a train from \textcolor{blue}{peterborough} on \textcolor{blue}{friday}\textcolor{red}{.}                                                          \\
Gold   & The user is looking for a train from peterborough on friday, which leaves at 16:00.                                   \\ \midrule
Error type & \textbf{Wrong slot} : The model mismatches slot template of the given information.                                                                                              \\ \midrule
Pattern    & The user is looking for a train for \_\_\_\_ people from \_\_\_\_ to \_\_\_\_ on \_\_\_\_, which leaves at \_\_\_\_.  \\
Summary    & The user is looking for a train for \textcolor{blue}{2} people from \textcolor{blue}{bishops stortford} to \textcolor{blue}{cambridge} on \textcolor{blue}{thursday}, which \textcolor{red}{arrives by} 18:30. \\
Gold   & The user is looking for a train for 2 people from bishops stortford to cambridge on thursday, which leaves at 18:30.  \\ \bottomrule
\end{tabular}
\end{adjustbox}
\caption{Three common error types of DS2. Dialogue id's of examples: MUL0603, SNG0271, PMUL4126.}
\label{tab:errors}
\end{table*}

\begin{table}[ht]
\centering
\begin{adjustbox}{width=0.39\textwidth}
\begin{tabular}{@{}lc@{}}
\toprule
\multicolumn{1}{c}{Model}          & \begin{tabular}[c]{@{}c@{}}Inference Time\\ Complexity\end{tabular} \\ \midrule
DSTReader~\cite{gao2019dstreader} & $O(k \tau)$                                                    \\
TRADE~\cite{wu2019transferable}    & $O(k \tau)$                                                    \\
COMER~\cite{ren2019comer}         & $O(k \tau)$                                                    \\
SOM-DST~\cite{kim2020somdst}      & $O(k \tau)$                                                    \\
T5-DST~\cite{lin2021t5dst}         & $O(k \tau)$                                                    \\
STARC~\cite{gao2020starc}          & $O(k \tau)$                                                    \\
TransferQA~\cite{lin2021zero}      & $O(k \tau)$                                                    \\
CINS~\cite{mi2021cins}             & $O(k \tau)$                                                    \\
PPTOD~\cite{su2021multi}           & $O(k + \tau)$                                                  \\
NADST~\cite{le2019nadst}           & $O(k + \tau)$                                                  \\
DS2 (Ours)                         & $O(k + \tau)$                                                  \\
\bottomrule
\end{tabular}
\end{adjustbox}
\caption{Worst-case inference time complexity adapted from ~\citet{ren2019comer,kim2020somdst}.
$k$ for number of slots and $\tau$ for model inference time.}
\label{tab:time_complexity}
\end{table}

\begin{table}[t]
\centering
\begin{adjustbox}{width=0.4\textwidth}
\begin{tabular}{@{}ll@{}}
\toprule
\textbf{Training Options}                           & JGA (std) \\ \midrule
DS2 (BART-large)                           & 28.3 (0.98)          \\ \midrule
- \textit{SAMSum pre-training}              & 25.5 (1.46)          \\ \midrule
- \textit{dontcare concat}                          & 27.1 (0.97)          \\
- \textit{paraphrasing}                    & 23.6 (0.71)          \\
- \textit{paraphrasing} \& \textit{dontcare concat} & 23.5 (1.86)          \\
- \textit{summary naturalness}             & 13.1 (0.45)          \\ \bottomrule
\end{tabular}
\end{adjustbox}
\caption{Effects of SAMSum pre-training and template naturalness.
Each row subtracts a module from the best setting of DS2. We show 3-run validation JGA for \MD~ 1\% few shot training of BART-large on 2.1.}
\label{tab:ablation}
\end{table}

\subsection{Few-shot: per-domain}
Table~\ref{tab:cd_jgas} shows the result of the few shot performance of DS2 compared to the baselines in three different settings described in Section~\ref{sec:few-shot-settings}. 
To compare with previous studies, we also evaluate our model on MultiWOZ 2.0 version. In \textit{ver.} 2.0, \citet{lin2021zero,gao2020starc} show that even without cross-domain pre-training \CT~ models can outperform \CD~ ones. We believe that this is can be attributed to the usage of large pre-trained language models like T5-large ($\sim$770M parameters). When we use the same-sized model, we outperform all other \CT~ models in \textbf{1\% setting} (30$\sim$50 dialogues) for 3 domains and achieve very competitive results in the 2 other domains.
When evaluating \textit{ver.} 2.0's SOTA model TransferQA on \textit{ver.} 2.1, we can, in fact, see that DS2 significantly outperforms it in all domains. We show slot accuracy and other metrics in Appendix Table~\ref{tab:rouge_table}.

\subsection{Few shot: all-domain}
In Table~\ref{tab:md_jgas}, we also show all-domain few-shot performance of DS2 in the \MD~setting compared to previous works. From the table, it is clear that for all 1\%, 5\%, 10\% few-shot adaptation settings, DS2 achieves SOTA performance in both MultiWOZ 2.0 and 2.1. It is also worth noting that we outperform PPTOD which not only uses T5-Large as well but also pre-trains their model on various TOD tasks and datasets. In addition, in the table, we also report the full-shot performance of DS2 which is 54.78 (2.0) \& 52.32 (2.1): relatively strong numbers considering that we did not put in any task-specific engineering as in ~\citet{heck2020trippy,yu2020score}.


\section{Analysis}

\subsection{Time Complexity}
Our method DS2 is efficient in terms of inference speed. Table~\ref{tab:time_complexity} shows inference time complexity. $k$ and $\tau$ each denote the number of slots and the model inference time. The numbers for other models are modified from~\citet{ren2019comer} and~\citet{kim2020somdst}. Other models, except for the bottom three models including DS2, has $O(k\tau)$ time complexity. For instance, QA-based models should ask a question for every potential slot in the given domains, so it requires $k$ times more model inference. On the other hand, DS2 only needs to run the PLM once for summary generation. After that, \textit{summary-to-state} pattern matching takes $O(k)$ time.

\subsection{Ablation Study}
\label{subsec:ablation_study}
In this section, we analyze the key components that to our model's success.

\paragraph{Dialogue Summary Pre-training}
As mentioned in Section~\ref{sec:model}, we further pre-train T5-large on the SAMSum corpus. The second row of Table~\ref{tab:ablation} shows what led to this decision. We observe that pre-training SAMSum had a large effect on BART-large ($\sim$440M parameters). In addition, we include the evaluation results on SAMSum in the Table~\ref{tab:dial_summ}; overall T5-large performed better than other models.

\paragraph{Summary Naturalness}
As mentioned in the last paragraph of Section~\ref{sec:methodology_overview}, guiding the generated summaries to conform to our synthetic templates is not a trivial task, and we hypothesized that the naturalness of these templates is key to successful performance. To answer this question, we conducted an ablation study on the \textit{state-to-summary} converter in Table~\ref{tab:ablation}.
The details of each \textit{state-to-summary} converter is shown in the Appendix Table~\ref{tab:converter_template_ex}.
In short, 1) \textit{paraphrasing} refers to whether we allow multiple prefixes and pronouns when synthesizing summary labels, 2) \textit{dontcare concat} is whether we use single or two sentences when adding \textit{dontcare} related phrases, and 3) finally, \textit{summary naturalness} is whether we use human-like language and grammar when making the summary. 
From the table, we can clearly see a significant performance drop when we disable \textit{summary naturalness}. Meanwhile, disabling \textit{paraphrasing} also had a non-negligible impact on the JGA, but \textit{dontcare concat} had only a minor decrease in performance.
Therefore, we conjecture that because we provide much more natural labels to the model, we can outperform PPTOD in Table~\ref{tab:md_jgas}.

\subsection{Error Analysis}
Table \ref{tab:errors} shows failure cases of DS2 summary model. Correctly predicted slot values are highlighted with blue color, while wrong ones are red. We report three categories of typical failures: ``hallucination'', ``missing slot'', and ``wrong slot''. \citet{Shuster2021RetrievalAR}, \citet{durmus2020feqa} reports ``Hallucination'' is a phenomenon in which a model generates unmentioned information in original dialogue. ``Missing slot'' is the most commonly observed case where the predicted summary omits information on a required slot. Similar failures also happen at the domain-level. The third type is named ``wrong slot'', where the model confuses two slots with same data types. For example, values for both ``arrive-by'' and ``leaves-at'' have same formats, so the model often fails to discriminating them. 



\section{Cost of Template Engineering}
For MultiWoZ, we devised templates for all 30 slots in the 5 domains that were used. Based on names of the slots, we wrote a \texttt{state-to-summary} function that generates a natural phrase along with the slot value with prefix templates. The \texttt{summary-to-state} parsing functions were written using regular expressions based on the rules we implemented for template generation. 
Overall, this process took approximately one week for one expert to finish. We believe that this is a much lower cost compared to full DST data design and collection efforts. Applying to a new domain may take even lower costs by using our code-base. Appendix Section~\ref{sec:guide} describes this process in detail.

\section{Limitations}
\label{sec:limitations}

In this section, we discuss several limitations of this work.
First, applying our model to a new domain requires a new summary template. Since DS2 performance is sensitive to the quality of the template as shown in the ablation study, considerable amount of knowledge on both domain and NLP is desired. However, following the guide in Section~\ref{sec:guide} would take less than one week for a researcher, which costs much less than collecting full DST data.
Second, DS2 is not capable of zero-shot inference because it should learn the template, at least from a few samples. 
Third, regular expression pattern matching may fail during the state extraction. There is no guarantee for the model output to fit in the template. The matching may still fail for a correctly formatted summary if a value entity contains template-like patterns. Using a neural network-based converter might easily solve this problem. 
Fourth, there is still room for improvement using DST-specific engineering (span matching or ontology searching as in TripPy~\cite{heck2020trippy}).
Finally, output summary length is bounded by the PLM's maximum sequence length, so DS2 might fail when we have too many slot values.
We leave these for future investigation.

\section{Conclusion}


This work tackles the few-shot DST problem by reformulating it into dialogue summarization. 
The strategy is to minimize the pre-train and fine-tune discrepancy by adapting a Pre-trained Language Model (PLM) to a more familiar task: summarization.
Hence, instead of forcing the model to learn a completely new task like DST, we provide rule-based summary templates from dialogue states.
We guide the summarization to conform to such templates, and utilize heuristic dialogue state extraction from the generated summaries.
The experimental results show that our model DS2 outperforms baselines for few-shot DST on MultiWoZ in both cross-domain and multi-domain settings. 
In addition, DS2 significantly reduces inference time complexity compared to existing QA-based methods. We also observed that naturalness of the template was very important.



\section*{Acknowledgements}

We would like to thank Whakyeong Seo and Wansoo Kim of Riiid very much for their gracious support on designing the figures and helping us scale up our experiments to the Google Cloud Platform. We would also like to thank Zhaonjiang Lin for the helpful discussions.
\bibliography{custom}
\bibliographystyle{acl_natbib}

\clearpage
\appendix

\section{Appendix}
\label{sec:appendix}


\subsection{\textit{State-to-summary} Ablation Details}

DST Performance improvement is driven by naturalness of summary template used for summary generation. To provide understanding of converting options we explored, Table~\ref{tab:domain_template_ex} shows an example sentence for each domain. In the table, all kinds of slots are introduced with example values. Example summary is constructed combining given slot values with corresponding slot templates as Table~\ref{tab:slot_template}. Last row shows the case of multi-domain dialogue. Sentence of each domain is concatenated by a conjunction `Also', in random order for balanced training.

Table~\ref{tab:converter_template_ex} shows difference between several converting options, which performance is compared in the previous Table~\ref{tab:ablation}. Unnatural converter was proposed to make a prediction without domain knowledge, so it generates slot names itself, while other converters do not generate names of domain or slot name.

Other ablation options are compared to each other under fair condition. The second option from the bottom was our initial idea. All domain's summary sentence shares same sentence prefix and the summary for \emph{dontcare} value was handled separately due to its quite different semantics from other values.
Paraphrasing seems to be effective, and we assume that is because the model was trained to avoid repetition of same phrase during their generative pre-train tasks. Concatenating \emph{dontcare} sentence is proposed from the idea that if the dialogue has several domain and more then one domain contains \emph{dontcare} slot, the number of sentence might be too large.

\subsection{Experiment Details}
We used cloud computing instances with NVIDIA Tesla A100 GPU for pre-training and fine-tuning t5-large model, and on-premise computing environment with NVIDIA GeForce RTX 2080 Ti for training BART-large model. Except pre-training task of \citet{lin2021zero} model as MultiWOZ 2.1 baseline implementation, we didn't use any distributed data parallel setting, so multiple GPUs are not used to train our DS2 model.

All experiments for DS2 used \texttt{pytorch-lightning} and \texttt{huggingface} libraries. Requirements for other software used are specified in the \texttt{requirements.txt} in the accompanying code. Training epoch is fixed with 100, while early stopping callback was enabled with patient 10 on validation joint goal accuracy metric, so most of the final number of epochs are within 10~30 epochs. Training batch size was 2 for T5-large, and 1 for BART. We used greedy search on transformer model's auto regressive language generation for speed, by setting number of beams parameter of pytorch model as 1. Accumulating gradient batches options are available in pytorch lightning trainer module, we set accumulate grad batches options to 1, 5, 10, 100 for 1\%, 5\%, 10\%, 100\% few shot learning. MultiWOZ dataset provides train, dev, test splits, so we used the given splits.

\begin{table}[t]
\begin{tabular}{@{}llll@{}}
\toprule
T5-Large            & \texttt{CT} & \texttt{CD} & \texttt{MD} \\ \midrule
1\%                 & 8$\sim$10     & 8$\sim$10    & 14$\sim$16           \\
5\%                 & 10$\sim$12    & 10$\sim$12   & 15$\sim$17           \\
10\%                & 12$\sim$14    & 12$\sim$14   & 16$\sim$18           \\
100\% / Pretraining & -          & 30$\sim$40   & 30$\sim$40   \\ \bottomrule
\end{tabular}
\caption{Estimated train/validation time (GPU hours) through virtual resource usage record. We used NVIDIA Tesla A100 through Google Cloud Platform.}
\label{tab:t5-time}
\end{table}

\subsection{Dataset and Model}

Table~\ref{tab:dialogue_count} shows number of dialogues in MultiWOZ 2.1 datasets. Single-domain dialogue is defined by a dialogue annotated with only one domain. Since appearance of unrelated domain information on dialogue may harm summarization's nature due to difference from dialogue state information to original text, single domain dialogues are the ideal requirements for cross-task setting. Lack of these single-domain dialogue as shown in table leads us to focus on cross domain setting, which can be performed naturally. 

Information we used on model selection is on Table~\ref{tab:dial_summ}. Comparing to previous study's reported performance of summarization on SAMSum corpus, T5-large does summarization well. Since larger models like T5-3B is harder to train with limited GPU resource, and previous work, \citet{lin2021zero} was also evaluated by T5-large, we selected T5-large as base summarization model. While there is no public T5-large model weight trained with SAMSum data, we pretrained T5-large by using the code from CODS\footnote{\url{https://github.com/salesforce/ConvSumm/tree/master/CODS}}.

In addition to T5-large, we also did many experiments using BART-large because smaller model weights of BART allows to be trained in single 2080Ti GPU, which costs much lower. For the comparative ablation experiment introduced in Table \ref{tab:ablation}, we used off-the-shelf weights for both SAMSum-unseen\footnote{\url{https://huggingface.co/facebook/bart-large-xsum}} and SAMSum-pretrained\footnote{\url{https://huggingface.co/Salesforce/bart-large-xsum-samsum}} weights.

\subsection{Guide for applying DS2 to a new domain}
\label{sec:guide}
The most plausible scenario of reusing our code is to apply it to a new dialogue domain. 
For that purpose, it is sufficient to rewrite the heuristic converters between dialogue states and summaries. It might take several hours for a Python developer to implement.

As explained in \ref{sec:state-to-sum}, our converting method is built in a hierarchical manner. Therefore, following the original design is the best strategy to add code for a new domain.


\begin{enumerate}
    \item Define natural language description for new domain.
        \begin{itemize}
            \item Define natural language templates for each domain and slot. e.g) in case of the slot "hotel-name", we can create a summary template sentence "The user is looking for a place to stay called {x}."
            \item Define natural language description for each slot to cover don't care scenario. e.g), in case of the slot "hotel-name", the summary sentence can be "The user is looking for a hotel and he does not care about the name".
        \end{itemize}
    \item Replace the code with your expression.
        \begin{itemize}
            \item We explicitly defined natural language expressions with python dictionary at the top of the converter python script. Inject your expression into the corresponding dictionary.
        \end{itemize}
    \item Modify converter code if you want to control plural form, article, space, or quotation marks. The final state-to-summary converter is written as Code~\ref{lst:state-to-sum}.
    \item Write a summary-to-state converter for the domain according to the intended expression as in Code~\ref{lst:sum-to-state}. 
\end{enumerate}

\definecolor{codegreen}{rgb}{0,0.6,0}
\definecolor{codegray}{rgb}{0.5,0.5,0.5}
\definecolor{codepurple}{rgb}{0.58,0,0.82}
\definecolor{backcolour}{rgb}{0.95,0.95,0.92}

\lstdefinestyle{mystyle}{
    backgroundcolor=\color{backcolour},   
    commentstyle=\color{codegreen},
    keywordstyle=\color{magenta},
    numberstyle=\tiny\color{codegray},
    stringstyle=\color{codepurple},
    basicstyle=\ttfamily\footnotesize,
    breakatwhitespace=false,         
    breaklines=true,
    captionpos=b,                    
    keepspaces=true,                 
    numbers=left,                    
    numbersep=5pt,                  
    showspaces=false,                
    showstringspaces=false,
    showtabs=false,                  
    tabsize=2
}

\lstset{style=mystyle}
\renewcommand{\lstlistingname}{Code}

\begin{table*}
\centering
\begin{lstlisting}[language=Python,label={lst:state-to-sum},caption=State-to-summary converter in Python code for the domain 'hotel'.
]
def hotel_state_to_sum(ds: dict, either: callable, is_one_sentence: bool, idx: int, wo_para: bool) -> str:
    first_sentence = get_first_sentence(ds=ds, domain="hotel", either=either, except_keys={"hotel-parking", "hotel-internet"}, idx=idx, wo_para=wo_para)

    second_sentence = get_dontcare_sentence(
        ds,
        domain="hotel",
        either=either,
        is_one_sentence=is_one_sentence,
        wo_para=wo_para
    )

    res = first_sentence + second_sentence + "."
    return res

\end{lstlisting}
\end{table*}

\begin{table*}
\centering
\begin{lstlisting}[language=Python,label={lst:sum-to-state},caption=Summary-to-state converter in Python code for the domain 'hotel'.
]
import re 
...

def hotel_sum_to_state(summ: str, is_one_sentence: bool) -> dict:
    sentences = re.split("|".join(COMMON_PHRASES), summ)
    summary = [sentence for sentence in sentences if DOMAIN_PHRASE_IN_SENTENCE["hotel"] in sentence]
    if not summary:
        return {}
    summary = summary[0]
    slot_to_prefix = {
        "hotel-type": " which is a ",
        "hotel-name": " called ",
        "hotel-stars": " ranked ",
        "hotel-pricerange": " with a",
        "hotel-area": " located in the ",
        "hotel-book people": r" for \d+ p",
        "hotel-book day": " on ",
        "hotel-book stay": r" for \d+ d",
        "hotel-parking": [" has no p", " has p"],
        "hotel-internet": [" has no i", " has i"],
    }
    res = {}

    dontcare_sentence = summary
    if not is_one_sentence:
        summary = summary.split('.')[0]

    for slot, prefix in slot_to_prefix.items():
        if type(prefix) == str:
            matches = [re.search(prefix, summary)]
        else:
            matches = [re.search(p, summary) for p in prefix]
        for match in matches:
            if match:
                start_idx = match.span()[-1]
                if slot in {"hotel-book people", "hotel-book stay"}:
                    start_idx -= 3
                elif slot == "hotel-pricerange":
                    start_idx += 2 if summary[start_idx:].startswith("n") else 1

                _summary = summary[start_idx:]

                value = re.split(
                    " The | Also, | which | called | ranked | during | located in the | for | on | and | with a| people| person| price| star| day",
                    _summary,
                )[0]

                if slot in ["hotel-internet", "hotel-parking"]:
                    value = "no" if " no " in match.group() else "yes"

                res[slot] = value.replace(",", "").replace(".", "")

    res.update(get_dontcare_values(dontcare_sentence, domain="hotel"))

    return res
\end{lstlisting}
\end{table*}

\begin{table*}[ht]
\centering
\begin{adjustbox}{width=1.0\linewidth}
\begin{tabular}{@{}lc@{}}
\toprule
\multicolumn{2}{c}{\textbf{ACL Reproducibility Guideline}}                                                 \\ \midrule
\multicolumn{2}{l}{\textbf{For all reported experimental results}}                                         \\ \midrule
A clear description of the mathematical setting, algorithm, and/or model                     & O           \\ \midrule
\begin{tabular}[c]{@{}l@{}}A link to (anonymized, for submission) downloadable source code, with specification of \\ all dependencies, including external libraries\end{tabular} &
  O \\ \midrule
A description of the computing infrastructure used                                           & O           \\ \midrule
The average runtime for each model or algorithm, or estimated energy cost                    & X           \\ \midrule
The number of parameters in each model                                                       & O           \\ \midrule
Corresponding validation performance for each reported test result                           & X           \\ \midrule
A clear definition of the specific evaluation measure or statistics used to report results   & O           \\ \midrule
\multicolumn{2}{l}{\textbf{For all results involving multiple experiments, such as hyperparameter search}} \\ \midrule
The exact number of training and evaluation runs                                             & O           \\ \midrule
The bounds for each hyperparameter                                                           & Not tuned   \\ \midrule
The hyperparameter configurations for best-performing models                                 & Not tuned   \\ \midrule
\begin{tabular}[c]{@{}l@{}}The method of choosing hyperparameter values (e.g., manual tuning, uniform sampling, etc.) \\ and the criterion used to select among them (e.g., accuracy)\end{tabular} &
  Not tuned \\ \midrule
Summary statistics of the results (e.g., mean, variance, error bars, etc.)                   & O           \\ \midrule
\multicolumn{2}{l}{\textbf{For all datasets used}}                                                         \\ \midrule
Relevant statistics such as number of examples and label distributions                       & O           \\ \midrule
Details of train/validation/test splits                                                      & O           \\ \midrule
An explanation of any data that were excluded, and all pre-processing steps                  & O           \\ \midrule
For natural language data, the name of the language(s)                                       & O           \\ \midrule
A link to a downloadable version of the dataset or simulation environment                    & O           \\ \midrule
\begin{tabular}[c]{@{}l@{}}For new data collected, a complete description of the data collection process, such as ownership\\  / licensing, informed consent, instructions to annotators and methods for quality control\end{tabular} &
  X \\ \bottomrule
\end{tabular}
\end{adjustbox}
\caption{Reproducibility Checklist. We do not do extensive hyper-parameter tuning for our models.}
\end{table*}

\begin{table*}[t]
\centering
\begin{adjustbox}{width=1.0\textwidth}
\begin{tabular}{@{}lccccc@{}}
\toprule
\multicolumn{1}{c}{\begin{tabular}[c]{@{}c@{}}T5-Large\\ Cross domain\end{tabular}} & JGA           & BLEU          & Slot True Acc & Slot None Acc & Rouge-4         \\ \midrule
taxi 1\%                                                                            & 0.764 (0.009) & 0.812 (0.008) & 0.729 (0.021) & 0.958 (0.003) & 0.797 (0.007) \\
taxi 5\%                                                                            & 0.798 (0.010) & 0.834 (0.008) & 0.769 (0.004) & 0.971 (0.005) & 0.820 (0.003)  \\
taxi 10\%                                                                           & 0.806 (0.004) & 0.839 (0.002) & 0.779 (0.005) & 0.973 (0.004) & 0.823 (0.003) \\ \midrule
hotel 1\%                                                                           & 0.430 (0.020) & 0.796 (0.008) & 0.810 (0.016)  & 0.939 (0.000)  & 0.785 (0.009) \\
hotel 5\%                                                                           & 0.484 (0.008) & 0.830 (0.002)  & 0.839 (0.005) & 0.955 (0.003) & 0.823 (0.002) \\
hotel 10\%                                                                          & 0.504 (0.011) & 0.836 (0.005) & 0.849 (0.011) & 0.957 (0.004) & 0.830 (0.004)  \\ \midrule
train 1\%                                                                           & 0.731 (0.008) & 0.828 (0.006) & 0.906 (0.004) & 0.972 (0.005) & 0.814 (0.006) \\
train 5\%                                                                           & 0.762 (0.004) & 0.860 (0.000)    & 0.917 (0.003) & 0.976 (0.003) & 0.843 (0.000)   \\
train 10\%                                                                          & 0.770 (0.005) & 0.863 (0.001) & 0.922 (0.002) & 0.977 (0.003) & 0.846 (0.001) \\ \midrule
attraction 1\%                                                                      & 0.600 (0.016) & 0.793 (0.006) & 0.761 (0.009) & 0.894 (0.013) & 0.773 (0.006) \\
attraction 5\%                                                                      & 0.687 (0.001) & 0.825 (0.006) & 0.840 (0.007)  & 0.909 (0.006) & 0.803 (0.006) \\
attraction 10\%                                                                     & 0.703 (0.004) & 0.832 (0.002) & 0.837 (0.002) & 0.927 (0.005) & 0.811 (0.002) \\ \midrule
restaurant 1\%                                                                      & 0.565 (0.031) & 0.811 (0.006) & 0.866 (0.037) & 0.941 (0.014) & 0.799 (0.007) \\
restaurant 5\%                                                                      & 0.651 (0.004) & 0.848 (0.001) & 0.907 (0.005) & 0.960 (0.001)  & 0.833 (0.001) \\
restaurant 10\%                                                                     & 0.673 (0.020) & 0.855 (0.004) & 0.910 (0.010)   & 0.962 (0.006) & 0.841 (0.004) \\ \bottomrule
\end{tabular}
\end{adjustbox}
\caption{Evaluation metrics for summary generation quality and slot prediction accuracy. Slot true accuracy means correctness rate for slots with existing value. Slot none accuracy is the metric for predict slots with none value as none. All of the value is mean (standard deviation) of three few shot trials}
\label{tab:rouge_table}
\end{table*}

\begin{table*}[t]
\begin{adjustbox}{width=\textwidth}
\begin{tabular}{@{}lll@{}}
\toprule
Domain &
  Example Dialogue State &
  Example summary \\ \midrule
Taxi &
  \begin{tabular}[c]{@{}l@{}}taxi-departure: \emph{london station}\\ taxi-destination: \emph{Incheon airport}\\ taxi-arriveby: \emph{12:30}\\ taxi-leaveat: \emph{02:45}\end{tabular} &
  \begin{tabular}[c]{@{}l@{}}The user is looking for a taxi from \emph{london station} \\ to \emph{Incheon airport}, which leaves at \emph{02:45} \\ and arrives by \emph{12:30}.\end{tabular} \\ \midrule
Train &
  \begin{tabular}[c]{@{}l@{}}train-departure: \emph{norwich}\\ train-destination: \emph{cambridge}\\ train-arriveby: \emph{19:45}\\ train-book people: \emph{3}\\ train-leaveat: \emph{11:21}\\ train-day: \emph{monday}\end{tabular} &
  \begin{tabular}[c]{@{}l@{}}The user is looking for a train for \emph{3} people from \\ \emph{norwich} to \emph{cambridge} on \emph{monday}, which \\ leaves at \emph{11:21} and arrives by \emph{19:45}.\end{tabular} \\ \midrule
Hotel &
  \begin{tabular}[c]{@{}l@{}}hotel-type: \emph{hotel}\\ hotel-name: \emph{Intercontinental}\\ hotel-stars: \emph{3}\\ hotel-pricerange: \emph{cheap}\\ hotel-area: \emph{east}\\ hotel-book people: \emph{6}\\ hotel-book day: \emph{saturday}\\ hotel-book stay: \emph{3}\\ hotel-parking: \emph{yes}\\ hotel-internet: \emph{no}\end{tabular} &
  \begin{tabular}[c]{@{}l@{}}The user is looking for a place to stay which is a \\ \emph{hotel} called \emph{Intercontinental} ranked \emph{3} stars with \\ a \emph{cheap} price located in the \emph{east} for \emph{6} people on \\ \emph{saturday} for \emph{3} days, which \emph{has parking} and \emph{has no internet}.\end{tabular} \\ \midrule
Attraction &
  \begin{tabular}[c]{@{}l@{}}attraction-area: \emph{cambridge}\\ attraction-name: \emph{nusha}\\ attraction-type: \emph{entertainment}\end{tabular} &
  \begin{tabular}[c]{@{}l@{}}The user is looking for an attraction which is \\ an \emph{entertainment} called \emph{nusha} located \\ in the \emph{cambridge}.\end{tabular} \\ \midrule
Restaurant &
  \begin{tabular}[c]{@{}l@{}}restaurant-book day: \emph{tuesday}\\ restaurant-book people: \emph{6}\\ restaurant-book time: \emph{12:00}\\ restaurant-name: \emph{meze bar}\\ restaurant-pricerange: \emph{cheap}\\ restaurant-area: \emph{south}\\ restaurant-food: \emph{seafood}\end{tabular} &
  \begin{tabular}[c]{@{}l@{}}The user is looking for a restaurant called \\ \emph{meze bar} located in the \emph{south} with a \emph{cheap} \\ price for \emph{6} people on \emph{tuesday} at \emph{12:00}, \\ which serves \emph{seafood}.\end{tabular} \\ \midrule
Multiple domain &
  \begin{tabular}[c]{@{}l@{}}restaurant-book day: \emph{tuesday}\\ restaurant-book time: \emph{12:00}\\ restaurant-name: \emph{meze bar}\\ train-departure: \emph{london station}\\ train-destination: \emph{Incheon airport}\\ train-book people: \emph{3}\\ hotel-type: \emph{guesthouse}\\ hotel-name: \emph{Intercontinental}\\ hotel-stars: \emph{3}\end{tabular} &
  \begin{tabular}[c]{@{}l@{}}The user is looking for a train for \emph{3} people from \\ \emph{london station} to \emph{Incheon airport}. Also, he is \\ searching for a restaurant called \emph{meze bar} on \\ \emph{tuesday} at \emph{12:00}. Also, he looks for a place to \\ stay which is a \emph{guesthouse} called \\ \emph{Intercontinental} ranked \emph{3} stars.\end{tabular} \\ \bottomrule
\end{tabular}
\end{adjustbox}
\caption{Example for summary template of each domain.}
\label{tab:domain_template_ex}
\end{table*}

\begin{table*}[ht]
\centering
\begin{adjustbox}{width=\textwidth}
\begin{tabular}{@{}lll@{}}
\toprule
\textbf{Sample Dialogue State} &
  \textbf{\begin{tabular}[c]{@{}l@{}}hotel-area: dontcare\\ hotel-pricerange: moderate\\ hotel-internet: yes\\ hotel-type: guesthouse\end{tabular}} &
  \textbf{\begin{tabular}[c]{@{}l@{}}train-book people: 3\\ train-leaveat: 10:30\\ train-destination: cambridge\\ train-day: tuesday\\ train-departure: kings lynn\end{tabular}} \\ \midrule
\textbf{Converter} &
  \multicolumn{2}{l}{\textbf{Example Summary}} \\ \midrule
Natural Summary (DS2) &
  \multicolumn{2}{l}{{\color[HTML]{333333} \begin{tabular}[c]{@{}l@{}}The user is looking for a place to stay which is a guesthouse with a moderate price,\\ which has internet\textcolor{blue}{, and he} does not care about the location. Also, \textcolor{blue}{he is searching}\\ \textcolor{blue}{for} a train for 3 people from kings lynn to cambridge on tuesday, which leaves at 10:30\end{tabular}}} \\ \midrule
\begin{tabular}[c]{@{}l@{}}Without paraphrasing \\ repeated prefix \\ (- \textit{paraphrasing})\end{tabular} &
  \multicolumn{2}{l}{{\color[HTML]{333333} \begin{tabular}[c]{@{}l@{}}The user is looking for a place to stay which is a guesthouse with a moderate price,\\ which has internet\textcolor{blue}{, and the user} does not care about the location. Also, \textcolor{blue}{the user} \\ \textcolor{blue}{is looking for} is looking for a train for 3 people from kings lynn to cambridge on \\tuesday, which leaves at 10:30.\end{tabular}}} \\ \midrule
\begin{tabular}[c]{@{}l@{}}Without concatenating \\ don't care sentence \\ (- \textit{dontcare concat})\end{tabular} &
  \multicolumn{2}{l}{\begin{tabular}[c]{@{}l@{}}The user is looking for a place to stay which is a guesthouse with a moderate price,\\ which has internet. \textcolor{blue}{He} does not care about the location. Also, \textcolor{blue}{he is searching for}\\ a train for 3 people from kings lynn to cambridge on tuesday, which leaves at 10:30.\end{tabular}} \\ \midrule
\begin{tabular}[c]{@{}l@{}}Without both paraphrasing, \\ concatenating\\  (- \textit{paraphrasing} \\ \& \textit{dontcare concat})\end{tabular} &
  \multicolumn{2}{l}{\begin{tabular}[c]{@{}l@{}}The user is looking for a place to stay which is a guesthouse with a moderate price,\\ which has internet. \textcolor{blue}{The user} does not care about the location. Also, \textcolor{blue}{the user is}\\ \textcolor{blue}{looking for} a train for 3 people from kings lynn to cambridge on tuesday, which\\ leaves at 10:30.\end{tabular}} \\ \midrule
\begin{tabular}[c]{@{}l@{}}Unnatural Summary \\ (- \textit{summary naturalness})\end{tabular} &
  \multicolumn{2}{l}{\begin{tabular}[c]{@{}l@{}}The user wants dontcare as area of hotel, moderate as pricerange of hotel, yes as\\ internet of hotel, guesthouse as type of hotel, 3 as book people of train, 10:30 as\\ leaveat of train, cambridge as destination of train, tuesday as day of train, kings\\ lynn as departure of train.\end{tabular}} \\ \bottomrule
\end{tabular}
\end{adjustbox}
\caption{Dialogue states from PMUL3853.json of MultiWOZ 2.1 and converted summary by using various converter options mentioned in Section 6.2. Differences by each converter options are pointed to \textcolor{blue}{blue} text color.}
\label{tab:converter_template_ex}
\end{table*}

\begin{table*}[ht]
\centering
\begin{adjustbox}{width=\textwidth}
\begin{tabular}{ccccccccccccccccc}
\hline
\multicolumn{2}{c}{\multirow{2}{*}{\begin{tabular}[c]{@{}c@{}}T5 Large\\ \textit{ver.} \& \texttt{mode}\end{tabular}}} &
  \multicolumn{3}{c}{Attraction} &
  \multicolumn{3}{c}{Hotel} &
  \multicolumn{3}{c}{Restaurant} &
  \multicolumn{3}{c}{Taxi} &
  \multicolumn{3}{c}{Train} \\ \cline{3-17} 
\multicolumn{2}{c}{} &
  1\% &
  5\% &
  10\% &
  1\% &
  5\% &
  10\% &
  1\% &
  5\% &
  10\% &
  1\% &
  5\% &
  10\% &
  1\% &
  5\% &
  10\% \\ \hline
\multirow{4}{*}{\begin{tabular}[c]{@{}c@{}}DS2\\ - 2.0\\ - \texttt{CD}\end{tabular}} &
  \begin{tabular}[c]{@{}c@{}}Run 1\\ (seed 11)\end{tabular} &
  65.79 &
  69.23 &
  73.34 &
  44.66 &
  52.09 &
  53.56 &
  59.63 &
  65.23 &
  66.33 &
  73.94 &
  77.42 &
  78.52 &
  75.05 &
  75.11 &
  77.58 \\
 &
  \begin{tabular}[c]{@{}c@{}}Run 2\\ (seed 23)\end{tabular} &
  65.76 &
  70.48 &
  70.35 &
  43.82 &
  53.37 &
  54.06 &
  57.52 &
  63.02 &
  63.53 &
  74.26 &
  76.52 &
  77.87 &
  72.40 &
  79.31 &
  80.21 \\
 &
  \begin{tabular}[c]{@{}c@{}}Run 3\\ (seed 47)\end{tabular} &
  64.24 &
  68.49 &
  68.97 &
  44.54 &
  51.03 &
  53.75 &
  59.66 &
  64.10 &
  64.10 &
  74.26 &
  77.61 &
  79.10 &
  75.16 &
  76.45 &
  78.00 \\
 &
  \begin{tabular}[c]{@{}c@{}}Mean\\ (Std.Dev)\end{tabular} &
  \begin{tabular}[c]{@{}c@{}}65.26\\ (0.89)\end{tabular} &
  \begin{tabular}[c]{@{}c@{}}69.40\\ (1.01)\end{tabular} &
  \begin{tabular}[c]{@{}c@{}}70.89\\ (2.23)\end{tabular} &
  \begin{tabular}[c]{@{}c@{}}44.34\\ (0.45)\end{tabular} &
  \begin{tabular}[c]{@{}c@{}}52.16\\ (1.17)\end{tabular} &
  \begin{tabular}[c]{@{}c@{}}53.79\\ (0.25)\end{tabular} &
  \begin{tabular}[c]{@{}c@{}}58.94\\ (1.23)\end{tabular} &
  \begin{tabular}[c]{@{}c@{}}64.12\\ (1.11)\end{tabular} &
  \begin{tabular}[c]{@{}c@{}}64.65\\ (1.48)\end{tabular} &
  \begin{tabular}[c]{@{}c@{}}74.15\\ (0.18)\end{tabular} &
  \begin{tabular}[c]{@{}c@{}}77.18\\ (0.58)\end{tabular} &
  \begin{tabular}[c]{@{}c@{}}78.50\\ (0.62)\end{tabular} &
  \begin{tabular}[c]{@{}c@{}}74.20\\ (1.56)\end{tabular} &
  \begin{tabular}[c]{@{}c@{}}76.96\\ (2.15)\end{tabular} &
  \begin{tabular}[c]{@{}c@{}}78.60\\ (1.41)\end{tabular} \\ \hline
\multirow{4}{*}{\begin{tabular}[c]{@{}c@{}}DS2\\ - 2.0\\ - \texttt{CT}\end{tabular}} &
  \begin{tabular}[c]{@{}c@{}}Run 1\\ (seed 11)\end{tabular} &
  56.82 &
  66.08 &
  70.71 &
  39.64 &
  48.88 &
  51.31 &
  50.49 &
  61.09 &
  65.11 &
  68.77 &
  72.32 &
  75.81 &
  70.24 &
  75.08 &
  79.05 \\
 &
  \begin{tabular}[c]{@{}c@{}}Run 2\\ (seed 23)\end{tabular} &
  56.01 &
  65.11 &
  68.62 &
  38.14 &
  47.03 &
  51.44 &
  44.54 &
  62.13 &
  65.11 &
  67.81 &
  72.84 &
  75.81 &
  68.74 &
  77.92 &
  78.84 \\
 &
  \begin{tabular}[c]{@{}c@{}}Run 3\\ (seed 47)\end{tabular} &
  54.69 &
  64.76 &
  66.85 &
  35.55 &
  48.16 &
  52.72 &
  50.67 &
  60.88 &
  63.62 &
  69.29 &
  72.65 &
  74.97 &
  72.13 &
  74.03 &
  76.58 \\
 &
  \begin{tabular}[c]{@{}c@{}}Mean\\ (Std.Dev)\end{tabular} &
  \begin{tabular}[c]{@{}c@{}}55.84\\ (1.08)\end{tabular} &
  \begin{tabular}[c]{@{}c@{}}65.32\\ (0.68)\end{tabular} &
  \begin{tabular}[c]{@{}c@{}}68.73\\ (1.93)\end{tabular} &
  \begin{tabular}[c]{@{}c@{}}37.78\\ (2.07)\end{tabular} &
  \begin{tabular}[c]{@{}c@{}}48.02\\ (0.93)\end{tabular} &
  \begin{tabular}[c]{@{}c@{}}51.82\\ (0.78)\end{tabular} &
  \begin{tabular}[c]{@{}c@{}}48.57\\ (3.49)\end{tabular} &
  \begin{tabular}[c]{@{}c@{}}61.37\\ (0.67)\end{tabular} &
  \begin{tabular}[c]{@{}c@{}}64.61\\ (0.86)\end{tabular} &
  \begin{tabular}[c]{@{}c@{}}68.62\\ (0.75)\end{tabular} &
  \begin{tabular}[c]{@{}c@{}}72.60\\ (0.26)\end{tabular} &
  \begin{tabular}[c]{@{}c@{}}75.53\\ (0.48)\end{tabular} &
  \begin{tabular}[c]{@{}c@{}}70.37\\ (1.70)\end{tabular} &
  \begin{tabular}[c]{@{}c@{}}75.68\\ (2.01)\end{tabular} &
  \begin{tabular}[c]{@{}c@{}}78.16\\ (1.37)\end{tabular} \\ \hline
\multirow{4}{*}{\begin{tabular}[c]{@{}c@{}}DS2\\ - 2.0\\ - \texttt{MD}\end{tabular}} &
  \begin{tabular}[c]{@{}c@{}}Run 1\\ (seed 11)\end{tabular} &
  63.70 &
  71.03 &
  70.32 &
  39.54 &
  51.59 &
  51.12 &
  52.60 &
  62.46 &
  64.75 &
  70.19 &
  75.68 &
  76.90 &
  69.98 &
  77.42 &
  78.39 \\
 &
  \begin{tabular}[c]{@{}c@{}}Run 2\\ (seed 23)\end{tabular} &
  61.93 &
  68.62 &
  70.93 &
  42.17 &
  52.15 &
  53.84 &
  55.40 &
  62.13 &
  65.14 &
  72.00 &
  75.29 &
  77.16 &
  71.27 &
  75.37 &
  76.24 \\
 &
  \begin{tabular}[c]{@{}c@{}}Run 3\\ (seed 47)\end{tabular} &
  61.22 &
  68.26 &
  71.38 &
  34.24 &
  48.10 &
  48.63 &
  55.37 &
  61.36 &
  63.68 &
  70.90 &
  74.32 &
  76.65 &
  69.98 &
  74.82 &
  79.60 \\
 &
  \begin{tabular}[c]{@{}c@{}}Mean\\ (Std.Dev)\end{tabular} &
  \begin{tabular}[c]{@{}c@{}}62.28\\ (1.28)\end{tabular} &
  \begin{tabular}[c]{@{}c@{}}69.30\\ (1.51)\end{tabular} &
  \begin{tabular}[c]{@{}c@{}}70.88\\ (0.53)\end{tabular} &
  \begin{tabular}[c]{@{}c@{}}38.65\\ (4.04)\end{tabular} &
  \begin{tabular}[c]{@{}c@{}}50.61\\ (2.19)\end{tabular} &
  \begin{tabular}[c]{@{}c@{}}51.20\\ (2.61)\end{tabular} &
  \begin{tabular}[c]{@{}c@{}}54.46\\ (1.61)\end{tabular} &
  \begin{tabular}[c]{@{}c@{}}61.98\\ (0.56)\end{tabular} &
  \begin{tabular}[c]{@{}c@{}}64.52\\ (0.76)\end{tabular} &
  \begin{tabular}[c]{@{}c@{}}71.03\\ (0.91)\end{tabular} &
  \begin{tabular}[c]{@{}c@{}}75.10\\ (0.70)\end{tabular} &
  \begin{tabular}[c]{@{}c@{}}76.90\\ (0.26)\end{tabular} &
  \begin{tabular}[c]{@{}c@{}}70.41\\ (0.74)\end{tabular} &
  \begin{tabular}[c]{@{}c@{}}75.87\\ (1.37)\end{tabular} &
  \begin{tabular}[c]{@{}c@{}}78.08\\ (1.70)\end{tabular} \\ \hline
\multirow{4}{*}{\begin{tabular}[c]{@{}c@{}}DS2\\ - 2.1\\ - \texttt{CD}\end{tabular}} &
  \begin{tabular}[c]{@{}c@{}}Run 1\\ (seed 11)\end{tabular} &
  57.88 &
  68.94 &
  70.45 &
  45.44 &
  47.82 &
  49.34 &
  59.04 &
  65.41 &
  68.62 &
  75.68 &
  80.19 &
  81.23 &
  71.92 &
  76.00 &
  77.37 \\
 &
  \begin{tabular}[c]{@{}c@{}}Run 2\\ (seed 23)\end{tabular} &
  61.58 &
  68.68 &
  69.74 &
  42.95 &
  48.00 &
  51.81 &
  58.41 &
  65.44 &
  68.68 &
  75.87 &
  78.39 &
  80.32 &
  73.87 &
  75.79 &
  77.37 \\
 &
  \begin{tabular}[c]{@{}c@{}}Run 3\\ (seed 47)\end{tabular} &
  60.68 &
  68.59 &
  70.74 &
  40.67 &
  49.50 &
  49.91 &
  52.16 &
  64.48 &
  64.48 &
  77.68 &
  80.84 &
  80.32 &
  73.42 &
  76.76 &
  76.26 \\
 &
  \begin{tabular}[c]{@{}c@{}}Mean\\ (Std.Dev)\end{tabular} &
  \begin{tabular}[c]{@{}c@{}}60.04\\ (1.93)\end{tabular} &
  \begin{tabular}[c]{@{}c@{}}68.74\\ (0.18)\end{tabular} &
  \begin{tabular}[c]{@{}c@{}}70.31\\ (0.51)\end{tabular} &
  \begin{tabular}[c]{@{}c@{}}43.02\\ (2.39)\end{tabular} &
  \begin{tabular}[c]{@{}c@{}}48.44\\ (0.92)\end{tabular} &
  \begin{tabular}[c]{@{}c@{}}50.35\\ (1.29)\end{tabular} &
  \begin{tabular}[c]{@{}c@{}}56.54\\ (3.80)\end{tabular} &
  \begin{tabular}[c]{@{}c@{}}65.11\\ (0.55)\end{tabular} &
  \begin{tabular}[c]{@{}c@{}}67.26\\ (2.41)\end{tabular} &
  \begin{tabular}[c]{@{}c@{}}76.41\\ (1.10)\end{tabular} &
  \begin{tabular}[c]{@{}c@{}}79.81\\ (1.27)\end{tabular} &
  \begin{tabular}[c]{@{}c@{}}80.62\\ (0.53)\end{tabular} &
  \begin{tabular}[c]{@{}c@{}}73.07\\ (1.02)\end{tabular} &
  \begin{tabular}[c]{@{}c@{}}76.18\\ (0.51)\end{tabular} &
  \begin{tabular}[c]{@{}c@{}}77.00\\ (0.64)\end{tabular} \\ \hline
\multirow{4}{*}{\begin{tabular}[c]{@{}c@{}}DS2\\ - 2.1\\ - \texttt{CT}\end{tabular}} &
  \begin{tabular}[c]{@{}c@{}}Run 1\\ (seed 11)\end{tabular} &
  52.64 &
  64.12 &
  67.40 &
  33.68 &
  46.97 &
  47.94 &
  45.79 &
  63.38 &
  66.66 &
  68.77 &
  75.55 &
  77.03 &
  64.54 &
  75.63 &
  77.79 \\
 &
  \begin{tabular}[c]{@{}c@{}}Run 2\\ (seed 23)\end{tabular} &
  51.77 &
  66.46 &
  67.65 &
  33.96 &
  48.06 &
  47.72 &
  47.96 &
  63.71 &
  65.76 &
  69.03 &
  76.45 &
  76.97 &
  68.8 &
  76.05 &
  76.66 \\
 &
  \begin{tabular}[c]{@{}c@{}}Run 3\\ (seed 47)\end{tabular} &
  56.40 &
  62.73 &
  65.66 &
  40.89 &
  45.85 &
  49.22 &
  51.32 &
  64.78 &
  68.03 &
  68.71 &
  78.45 &
  77.68 &
  70.53 &
  74.97 &
  76.97 \\
 &
  \begin{tabular}[c]{@{}c@{}}Mean\\ (Std.Dev)\end{tabular} &
  \begin{tabular}[c]{@{}c@{}}53.60\\ (2.46)\end{tabular} &
  \begin{tabular}[c]{@{}c@{}}64.44\\ (1.89)\end{tabular} &
  \begin{tabular}[c]{@{}c@{}}66.90\\ (1.08)\end{tabular} &
  \begin{tabular}[c]{@{}c@{}}36.18\\ (4.08)\end{tabular} &
  \begin{tabular}[c]{@{}c@{}}46.96\\ (1.11)\end{tabular} &
  \begin{tabular}[c]{@{}c@{}}48.29\\ (0.81)\end{tabular} &
  \begin{tabular}[c]{@{}c@{}}48.36\\ (2.79)\end{tabular} &
  \begin{tabular}[c]{@{}c@{}}63.96\\ (0.73)\end{tabular} &
  \begin{tabular}[c]{@{}c@{}}66.82\\ (1.14)\end{tabular} &
  \begin{tabular}[c]{@{}c@{}}68.84\\ (0.17)\end{tabular} &
  \begin{tabular}[c]{@{}c@{}}76.82\\ (1.48)\end{tabular} &
  \begin{tabular}[c]{@{}c@{}}77.23\\ (0.39)\end{tabular} &
  \begin{tabular}[c]{@{}c@{}}67.96\\ (3.08)\end{tabular} &
  \begin{tabular}[c]{@{}c@{}}75.55\\ (0.54)\end{tabular} &
  \begin{tabular}[c]{@{}c@{}}77.14\\ (0.58)\end{tabular} \\ \hline
\multirow{4}{*}{\begin{tabular}[c]{@{}c@{}}DS2\\ - 2.1\\ - \texttt{MD}\end{tabular}} &
  \begin{tabular}[c]{@{}c@{}}Run 1\\ (seed 11)\end{tabular} &
  55.34 &
  65.66 &
  67.75 &
  37.55 &
  47.66 &
  48.97 &
  48.02 &
  60.58 &
  62.43 &
  69.16 &
  76.19 &
  78.52 &
  68.77 &
  75.74 &
  75.29 \\
 &
  \begin{tabular}[c]{@{}c@{}}Run 2\\ (seed 23)\end{tabular} &
  56.33 &
  64.08 &
  67.49 &
  38.98 &
  47.60 &
  47.85 &
  50.43 &
  64.39 &
  65.64 &
  73.55 &
  76.65 &
  80.06 &
  71.32 &
  75.74 &
  76.89 \\
 &
  \begin{tabular}[c]{@{}c@{}}Run 3\\ (seed 47)\end{tabular} &
  57.33 &
  69.42 &
  66.17 &
  38.14 &
  48.00 &
  48.19 &
  52.13 &
  64.69 &
  65.29 &
  72.90 &
  78.45 &
  78.45 &
  69.51 &
  75.18 &
  76.89 \\
 &
  \begin{tabular}[c]{@{}c@{}}Mean\\ (Std.Dev)\end{tabular} &
  \begin{tabular}[c]{@{}c@{}}56.33\\ (1.00)\end{tabular} &
  \begin{tabular}[c]{@{}c@{}}66.39\\ (2.74)\end{tabular} &
  \begin{tabular}[c]{@{}c@{}}67.14\\ (0.85)\end{tabular} &
  \begin{tabular}[c]{@{}c@{}}38.22\\ (0.72)\end{tabular} &
  \begin{tabular}[c]{@{}c@{}}47.75\\ (0.22)\end{tabular} &
  \begin{tabular}[c]{@{}c@{}}48.34\\ (0.57)\end{tabular} &
  \begin{tabular}[c]{@{}c@{}}50.19\\ (2.07)\end{tabular} &
  \begin{tabular}[c]{@{}c@{}}63.22\\ (2.29)\end{tabular} &
  \begin{tabular}[c]{@{}c@{}}64.45\\ (1.76)\end{tabular} &
  \begin{tabular}[c]{@{}c@{}}71.87\\ (2.37)\end{tabular} &
  \begin{tabular}[c]{@{}c@{}}77.10\\ (1.19)\end{tabular} &
  \begin{tabular}[c]{@{}c@{}}79.01\\ (0.91)\end{tabular} &
  \begin{tabular}[c]{@{}c@{}}69.87\\ (1.31)\end{tabular} &
  \begin{tabular}[c]{@{}c@{}}75.55\\ (0.32)\end{tabular} &
  \begin{tabular}[c]{@{}c@{}}76.36\\ (0.92)\end{tabular} \\ \hline
\multirow{4}{*}{\begin{tabular}[c]{@{}c@{}}TransferQA\\ - 2.1\\ - \texttt{CT}\end{tabular}} &
  \begin{tabular}[c]{@{}c@{}}Run 1\\ (seed 577)\end{tabular} &
  48.94 &
  60.87 &
  65.34 &
  31.93 &
  38.95 &
  41.35 &
  49.75 &
  59.84 &
  62.82 &
  70.77 &
  74.52 &
  75.74 &
  68.95 &
  72.58 &
  75.95 \\
 &
  \begin{tabular}[c]{@{}c@{}}Run 2\\ (seed 17)\end{tabular} &
  50.03 &
  61.38 &
  62.89 &
  34.21 &
  38.76 &
  40.79 &
  45.01 &
  60.73 &
  61.98 &
  74.13 &
  73.42 &
  76.52 &
  69.77 &
  73.61 &
  75.03 \\
 &
  \begin{tabular}[c]{@{}c@{}}Run 3\\ (seed 117)\end{tabular} &
  51.77 &
  60.51 &
  64.60 &
  31.24 &
  39.36 &
  43.82 &
  46.59 &
  56.92 &
  61.92 &
  68.45 &
  75.48 &
  75.94 &
  68.32 &
  73.32 &
  75.39 \\
 &
  \begin{tabular}[c]{@{}c@{}}Mean\\ (Std.Dev)\end{tabular} &
  \begin{tabular}[c]{@{}c@{}}50.25\\ (1.43)\end{tabular} &
  \begin{tabular}[c]{@{}c@{}}60.92\\ (0.44)\end{tabular} &
  \begin{tabular}[c]{@{}c@{}}64.28\\ (1.26)\end{tabular} &
  \begin{tabular}[c]{@{}c@{}}32.46\\ (1.55)\end{tabular} &
  \begin{tabular}[c]{@{}c@{}}39.02\\ (0.31)\end{tabular} &
  \begin{tabular}[c]{@{}c@{}}41.99\\ (1.61)\end{tabular} &
  \begin{tabular}[c]{@{}c@{}}47.12\\ (2.41)\end{tabular} &
  \begin{tabular}[c]{@{}c@{}}59.16\\ (1.99)\end{tabular} &
  \begin{tabular}[c]{@{}c@{}}62.24\\ (0.50)\end{tabular} &
  \begin{tabular}[c]{@{}c@{}}71.12\\ (2.86)\end{tabular} &
  \begin{tabular}[c]{@{}c@{}}74.47\\ (1.03)\end{tabular} &
  \begin{tabular}[c]{@{}c@{}}76.07\\ (0.41)\end{tabular} &
  \begin{tabular}[c]{@{}c@{}}69.01\\ (0.73)\end{tabular} &
  \begin{tabular}[c]{@{}c@{}}73.17\\ (0.53)\end{tabular} &
  \begin{tabular}[c]{@{}c@{}}75.46\\ (0.46)\end{tabular} \\ \hline
\end{tabular}
\end{adjustbox}
\caption{Few-shot (1-5-10\%) results on MultiWoZ 2.0 and 2.1 (\textit{ver.}). \texttt{CD}, \texttt{CT}, \texttt{MD} each refer to \textit{Cross-Domain}, \textit{Cross-Task}, \textit{Multi-Domain} few-shot scenarios. Full results and statistics of each run are provided here.}
\label{tab:all_run_jgas}
\end{table*}

\begin{table*}[t]
\centering
\begin{adjustbox}{width=\textwidth}
\begin{tabular}{@{}ccccccccccccccccc@{}}
\toprule
\multicolumn{2}{c}{\multirow{2}{*}{\begin{tabular}[c]{@{}c@{}}BART-Large\\ \textit{ver.} \& \texttt{mode}\end{tabular}}} &
  \multicolumn{3}{c}{Attraction} &
  \multicolumn{3}{c}{Hotel} &
  \multicolumn{3}{c}{Restaurant} &
  \multicolumn{3}{c}{Taxi} &
  \multicolumn{3}{c}{Train} \\ \cmidrule(l){3-17} 
\multicolumn{2}{c}{} &
  1\% &
  5\% &
  10\% &
  1\% &
  5\% &
  10\% &
  1\% &
  5\% &
  10\% &
  1\% &
  5\% &
  10\% &
  1\% &
  5\% &
  10\% \\ \midrule
\multirow{4}{*}{\begin{tabular}[c]{@{}c@{}}DS2\\ - 2.1\\ - \texttt{CD}\end{tabular}} &
  \begin{tabular}[c]{@{}c@{}}Run 1\\ (seed 11)\end{tabular} &
  53.15 &
  62.51 &
  63.79 &
  33.99 &
  45.51 &
  49.22 &
  46.95 &
  59.66 &
  63.32 &
  68.58 &
  76.52 &
  79.55 &
  56.68 &
  73.69 &
  74.89 \\
 &
  \begin{tabular}[c]{@{}c@{}}Run 2\\ (seed 23)\end{tabular} &
  51.51 &
  62.80 &
  61.83 &
  34.33 &
  46.60 &
  48.47 &
  48.35 &
  61.45 &
  62.19 &
  68.26 &
  77.81 &
  79.10 &
  62.12 &
  73.00 &
  76.76 \\
 &
  \begin{tabular}[c]{@{}c@{}}Run 3\\ (seed 47)\end{tabular} &
  55.50 &
  65.59 &
  60.16 &
  34.80 &
  46.22 &
  47.94 &
  50.58 &
  61.66 &
  64.45 &
  69.23 &
  76.84 &
  80.84 &
  63.09 &
  73.13 &
  76.74 \\
 &
  \begin{tabular}[c]{@{}c@{}}Mean\\ (Std.Dev)\end{tabular} &
  \begin{tabular}[c]{@{}c@{}}53.39\\ (2.01)\end{tabular} &
  \begin{tabular}[c]{@{}c@{}}63.63\\ (1.70)\end{tabular} &
  \begin{tabular}[c]{@{}c@{}}61.93\\ (1.82)\end{tabular} &
  \begin{tabular}[c]{@{}c@{}}34.37\\ (0.41)\end{tabular} &
  \begin{tabular}[c]{@{}c@{}}46.11\\ (0.55)\end{tabular} &
  \begin{tabular}[c]{@{}c@{}}48.54\\ (0.64)\end{tabular} &
  \begin{tabular}[c]{@{}c@{}}48.63\\ (1.83)\end{tabular} &
  \begin{tabular}[c]{@{}c@{}}60.92\\ (1.10)\end{tabular} &
  \begin{tabular}[c]{@{}c@{}}63.32\\ (1.13)\end{tabular} &
  \begin{tabular}[c]{@{}c@{}}68.69\\ (0.49)\end{tabular} &
  \begin{tabular}[c]{@{}c@{}}77.06\\ (0.67)\end{tabular} &
  \begin{tabular}[c]{@{}c@{}}79.83\\ (0.90)\end{tabular} &
  \begin{tabular}[c]{@{}c@{}}60.63\\ (3.46)\end{tabular} &
  \begin{tabular}[c]{@{}c@{}}73.27\\ (0.37)\end{tabular} &
  \begin{tabular}[c]{@{}c@{}}76.13\\ (1.07)\end{tabular} \\ \midrule
\multirow{4}{*}{\begin{tabular}[c]{@{}c@{}}DS2\\ - 2.1\\ - \texttt{CT}\end{tabular}} &
  \begin{tabular}[c]{@{}c@{}}Run 1\\ (seed 11)\end{tabular} &
  39.87 &
  61.61 &
  64.50 &
  29.93 &
  42.63 &
  46.72 &
  37.30 &
  56.77 &
  62.31 &
  64.39 &
  60.92 &
  73.94 &
  56.28 &
  70.45 &
  75.81 \\
 &
  \begin{tabular}[c]{@{}c@{}}Run 2\\ (seed 23)\end{tabular} &
  39.20 &
  61.70 &
  59.74 &
  32.49 &
  44.07 &
  46.16 &
  39.77 &
  59.90 &
  59.81 &
  61.74 &
  63.32 &
  76.06 &
  64.17 &
  69.58 &
  74.00 \\
 &
  \begin{tabular}[c]{@{}c@{}}Run 3\\ (seed 47)\end{tabular} &
  41.41 &
  58.07 &
  60.68 &
  29.93 &
  41.04 &
  45.47 &
  37.45 &
  56.59 &
  62.01 &
  63.23 &
  70.00 &
  75.29 &
  46.90 &
  72.98 &
  73.79 \\
 &
  \begin{tabular}[c]{@{}c@{}}Mean\\ (Std.Dev)\end{tabular} &
  \begin{tabular}[c]{@{}c@{}}40.16\\ (1.13)\end{tabular} &
  \begin{tabular}[c]{@{}c@{}}60.46\\ (2.07)\end{tabular} &
  \begin{tabular}[c]{@{}c@{}}61.64\\ (2.52)\end{tabular} &
  \begin{tabular}[c]{@{}c@{}}30.78\\ (1.48)\end{tabular} &
  \begin{tabular}[c]{@{}c@{}}42.58\\ (1.52)\end{tabular} &
  \begin{tabular}[c]{@{}c@{}}46.12\\ (0.63)\end{tabular} &
  \begin{tabular}[c]{@{}c@{}}38.17\\ (1.38)\end{tabular} &
  \begin{tabular}[c]{@{}c@{}}57.75\\ (1.86)\end{tabular} &
  \begin{tabular}[c]{@{}c@{}}61.38\\ (1.37)\end{tabular} &
  \begin{tabular}[c]{@{}c@{}}63.12\\ (1.33)\end{tabular} &
  \begin{tabular}[c]{@{}c@{}}71.27\\ (2.54)\end{tabular} &
  \begin{tabular}[c]{@{}c@{}}75.10\\ (1.07)\end{tabular} &
  \begin{tabular}[c]{@{}c@{}}55.78\\ (8.65)\end{tabular} &
  \begin{tabular}[c]{@{}c@{}}71.00\\ (1.77)\end{tabular} &
  \begin{tabular}[c]{@{}c@{}}74.53\\ (1.11)\end{tabular} \\ \midrule
\multirow{4}{*}{\begin{tabular}[c]{@{}c@{}}DS2\\ - 2.1\\ - \texttt{MD}\end{tabular}} &
  \begin{tabular}[c]{@{}c@{}}Run 1\\ (seed 11)\end{tabular} &
  42.06 &
  61.32 &
  58.14 &
  30.49 &
  38.92 &
  45.13 &
  38.58 &
  51.83 &
  61.51 &
  61.16 &
  65.74 &
  66.65 &
  54.31 &
  72.95 &
  68.35 \\
 &
  \begin{tabular}[c]{@{}c@{}}Run 2\\ (seed 23)\end{tabular} &
  45.92 &
  53.83 &
  60.80 &
  33.40 &
  39.76 &
  43.29 &
  36.53 &
  56.30 &
  59.30 &
  59.94 &
  65.48 &
  71.68 &
  58.28 &
  68.09 &
  68.56 \\
 &
  \begin{tabular}[c]{@{}c@{}}Run 3\\ (seed 47)\end{tabular} &
  41.03 &
  55.40 &
  56.66 &
  32.62 &
  41.92 &
  47.75 &
  39.60 &
  62.10 &
  53.32 &
  60.77 &
  65.23 &
  67.81 &
  60.91 &
  68.85 &
  72.29 \\
 &
  \begin{tabular}[c]{@{}c@{}}Mean\\ (Std.Dev)\end{tabular} &
  \begin{tabular}[c]{@{}c@{}}43.00\\ (2.58)\end{tabular} &
  \begin{tabular}[c]{@{}c@{}}56.85\\ (3.95)\end{tabular} &
  \begin{tabular}[c]{@{}c@{}}58.53\\ (2.10)\end{tabular} &
  \begin{tabular}[c]{@{}c@{}}32.17\\ (1.51)\end{tabular} &
  \begin{tabular}[c]{@{}c@{}}40.20\\ (1.55)\end{tabular} &
  \begin{tabular}[c]{@{}c@{}}45.39\\ (2.24)\end{tabular} &
  \begin{tabular}[c]{@{}c@{}}38.24\\ (1.56)\end{tabular} &
  \begin{tabular}[c]{@{}c@{}}56.74\\ (5.15)\end{tabular} &
  \begin{tabular}[c]{@{}c@{}}58.04\\ (4.24)\end{tabular} &
  \begin{tabular}[c]{@{}c@{}}60.62\\ (0.62)\end{tabular} &
  \begin{tabular}[c]{@{}c@{}}65.48\\ (0.26)\end{tabular} &
  \begin{tabular}[c]{@{}c@{}}68.71\\ (2.63)\end{tabular} &
  \begin{tabular}[c]{@{}c@{}}57.83\\ (3.32)\end{tabular} &
  \begin{tabular}[c]{@{}c@{}}69.96\\ (2.61)\end{tabular} &
  \begin{tabular}[c]{@{}c@{}}69.73\\ (2.22)\end{tabular} \\ \bottomrule
\end{tabular}
\end{adjustbox}
\caption{Few-shot(1-5-10\%) results on MultiWoZ 2.1 with BART-Large model. Meaning of the fields are same as in Table \ref{tab:all_run_jgas}.}
\label{tab:bart_performance}
\end{table*}

\begin{table*}[t]
\centering
\begin{adjustbox}{width=0.7\textwidth}
\begin{tabular}{ccccc}
\hline
\multicolumn{2}{c}{\textbf{Few-shot ratio}}                  & \textbf{1\%} & \textbf{5\%} & \textbf{10\%} \\ \hline
\multirow{4}{*}{\textbf{DS2 - T5 (2.0)}}   & Run 1 (seed 11) & 35.67        & 46.21        & 47.86         \\
                                           & Run 2 (seed 23) & 38.22        & 46.01        & 47.79         \\
                                           & Run 3 (seed 47) & 34.57        & 43.19        & 47.18         \\
                                           & Mean (Std. Dev) & 36.15 (1.87) & 45.14 (1.69) & 47.61 (0.37)  \\ \hline
\multirow{4}{*}{\textbf{DS2 - T5 (2.1)}}   & Run 1 (seed 11) & 32.04        & 43.30        & 44.30         \\
                                           & Run 2 (seed 23) & 34.74        & 44.06        & 46.40         \\
                                           & Run 3 (seed 47) & 34.50        & 45.24        & 45.43         \\
                                           & Mean (Std. Dev) & 33.76 (1.49) & 44.2 (0.98)  & 45.38 (1.05)  \\ \hline
\multirow{4}{*}{\textbf{DS2 - BART (2.1)}} & Run 1 (seed 11) & 27.52        & 37.39        & 40.05         \\
                                           & Run 2 (seed 23) & 27.86        & 36.86        & 40.61         \\
                                           & Run 3 (seed 47) & 29.37        & 38.88        & 40.21         \\
                                           & Mean (Std. Dev) & 28.25 (0.98) & 37.71 (1.05) & 40.29 (0.29)  \\ \hline
\end{tabular}
\end{adjustbox}
\caption{Few-shot(1-5-10\%) all-domain results on MultiWoZ 2.0 \& 2.1 for multi-domain setting.}
\label{tab:md_all_performance}
\end{table*}

\end{document}